\documentclass[10pt,twocolumn,letterpaper]{article}

\usepackage{cvpr}
\usepackage{times}
\usepackage{epsfig}
\usepackage{subfig}
\usepackage{graphicx}
\usepackage{amsmath}
\usepackage{amssymb}
\usepackage{multirow}
\usepackage{todonotes}
\usepackage{dsfont}
\usepackage[ruled,vlined,linesnumbered]{algorithm2e}
\newcommand{\nxm}[2]{${#1}\times{#2}$}
\newcommand{\nxn}[1]{\nxm{#1}{#1}}

\newcommand{\figref}[2][]{\hyperref[#2]{Fig.\ \ref*{#2}{#1}}}
\renewcommand{\eqref}[1]{\hyperref[#1]{Eq.\ (\ref*{#1})}}
\newcommand{\secref}[1]{\hyperref[#1]{Section\ \ref*{#1}}}
\newcommand{\tabref}[1]{\hyperref[#1]{Table\ \ref*{#1}}}
\newcommand{\appref}[1]{\hyperref[#1]{Appendix\ \ref*{#1}}}

\usepackage{nohyperref}  % This makes hyperref commands do nothing, without errors
\usepackage{url}

\cvprfinalcopy % *** Uncomment this line for the final submission

\def\cvprPaperID{4951} % *** Enter the CVPR Paper ID here

\begin{document}

%%%%%%%%% TITLE
\title{Structured Pruning of Neural Networks with Budget-Aware Regularization \vspace{-0.3cm}}

\author{Carl Lemaire\\
Universit\'e de Sherbrooke\\
Sherbrooke, Canada\\
{\tt\footnotesize carl.lemaire@usherbrooke.ca}
\and
Andrew Achkar \\
Miovision Technologies Inc.\\
Kitchener, Canada\\
{\tt\footnotesize aachkar@miovision.com}
\and
Pierre-Marc Jodoin \\
Universit\'e de Sherbrooke\\
Sherbrooke, Canada\\
{\tt\footnotesize pierre-marc.jodoin@usherbrooke.ca}
}

\maketitle
%\thispagestyle{empty}

%%%%%%%%% ABSTRACT
\begin{abstract}
\vspace{-0.2cm}
   Pruning methods have shown to be effective at reducing the size of deep neural networks while keeping accuracy almost intact. Among the most effective methods are those that prune a network while training it with a sparsity prior loss and learnable dropout parameters.  A shortcoming of these approaches however is that neither the size nor the inference speed of the pruned network can be controlled directly; yet this is a key feature for targeting deployment of CNNs on low-power hardware. To overcome this, we introduce a budgeted regularized pruning framework for deep CNNs. Our approach naturally fits into traditional neural network training as it consists of a learnable masking layer, a novel budget-aware objective function, and the use of knowledge distillation.  We also provide insights on how to prune a residual network and how this  can lead to new architectures.  Experimental results reveal that CNNs pruned with our method are more accurate and less compute-hungry than  state-of-the-art methods.  Also, our approach is more effective at preventing accuracy collapse in case of severe pruning; this allows pruning factors of up to $16\times$  without significant accuracy drop. \footnote{Our code is available here: https://tinyurl.com/lemaire2019} \vspace{-0.6cm}
   % TODO1: "A shortcoming of these approaches however is that neither the size nor the inference speed of the pruned network can be controlled directly": notre méthode ne permet pas non plus de contrôler directement ces métriques
   
   % We release the source code of our method and experiments for the sake of reproducibility. 
\end{abstract}

\section{Introduction}
\vspace{-0.2cm}
Convolutional Neural Networks (CNN) have proven to be effective feature extractors for many computer vision tasks \cite{he2017mask,huang2017densely, krizhevsky2012imagenet, ronneberger2015u}. The design of several CNNs involve many heuristics, such as using increasing powers of two as the number of feature maps, or \textit{width}, of each layer. While such heuristics allow achieving excellent results, they may be too crude in situations where the amount of compute power and memory is restricted, such as with mobile platforms. Thus arises the problem of finding the right number of layers that solve a given task while respecting a budget. Since the number of layers depends highly on the effectiveness of the learned filters (and their combination), one cannot determine these hyper-parameters {\em a priori}.

Convolution operations constitute the main computational burden of a CNN. The execution of these operations benefit from a high degree of parallelism, which requires them to have regular structures. This implies that one cannot remove isolated neurons from a CNN filter as they must be full grids. To achieve the same effect as removing a neuron, one can \textit{zero-out} its weights.  While doing this reduces the theoretical size of the model, it does not reduce the computational demands of the model nor the amount of  feature map memory.  Therefore, to accelerate a CNN and reduce its memory footprint, one has to rely on {\em structured} sparsity pruning methods that aim at reducing the number of feature maps and not just individual neurons.

% TODO "recover performance" -> other citations
By removing unimportant filters from a network and retraining it, one can shrink it while maintaining good performance~\cite{han2015learning, lecun1990optimal}. This can be explained by the following hypothesis: the initial value of a filter's weights is not guaranteed to allow the learning of a useful feature; thus, a trained network might contain many expendable features~\cite{frankle2018lottery}. %By identifying and removing these filters, it is possible to obtain a smaller network that still performs well. %[TODO: mentionner l'histoire du train from scratch $>$ pruning]

Among the structured pruning methods, those that implement a sparsity learning (SL) framework have shown to be effective as pruning and training are done simultaneously~\cite{dai2018infobottlen, kingma2015variational, Liu2017,louizos2018learning, molchanov2017variational, neklyudov2017structured}.  Unfortunately, most SL methods cannot prune a network while respecting a \textit{neuron budget} imposed by the very nature of a device on which the network shall be deployed.  As of today, 
% TODO1: Notre méthode ne permet pas de respecter un tel budget. La métrique que l'on a choisi: "activation tensor volume" n'est pas directement reliée à la quantité de mémoire requise. Il faudrait faire de l'essai-erreur quand même, mais l'essai-erreur serait probablement beaucoup plus rapide.
%The main interface for controlling the size of the network is a \textit{size-accuracy trade-off} hyper-parameter. If we want to aim for a specific number of FLOPS per inference, this hyper-parameter is not helpful; to set it, a 
pruning a network while respecting a budget can only be done by trial-and-error, typically by training multiple times a network with various pruning hyperparameters.  

In this paper, we present a SL framework which allows learning and selecting filters of a CNN while respecting a neuron budget.   Our main contributions are: \vspace{-0.2cm}

\begin{itemize}
    \item We present a novel objective function which includes a variant of the log-barrier~\cite{Boyd2004} function for simultaneously training and pruning a CNN while respecting a total neuron budget;\vspace{-0.2cm}
    \item We propose a variant of the barrier method~\cite{Boyd2004} for optimizing a CNN;\vspace{-0.2cm}
    \item We demonstrate the effectiveness of combining SL and knowledge distillation~\cite{hinton2015distilling};\vspace{-0.2cm}
    \item We empirically confirm the existence of the \textit{automatic depth determination} property of residual networks pruned with filter-wise methods, and give insights on how to ensure the viability of the pruned network by preventing ``fatal pruning'';  \vspace{-0.2cm}
    \item We propose a new {\em mixed-connectivity block} which roughly doubles the effective pruning factors attainable with our method.
\end{itemize}
\vspace{-0.4cm}
\section{Previous Works}\vspace{-0.2cm}

Compressing neural networks without affecting too much their accuracy implies that networks are often over-parametrized.  % \textbf{Neural network over-parametrization} is a widely studied subject. 
Denil \etal~\cite{denil2013predicting} have shown that typical neural networks are over-parametrized; in the best case of their experiments, they could predict 95\% of the network weights from the others.  Recent work by Frankle \etal~\cite{frankle2018lottery} support the hypothesis that a large proportion (typically 90\%) of weights in standard neural networks are initialized to a value that will lead to an expendable feature. %These results suggest that networks an order of magnitude smaller could exhibit similar performances than their larger counterparts, while fitting on resource-constrained devices. 
In this section, we review six categories of methods for reducing the size of a neural network.  %[TODO: talk about contradictions to frankle2018lottery]

% han2015learning,han2015deep
\textbf{Neural network compression} aims to reduce the storage requirements of the network's weights. In~\cite{denton2014exploiting,jaderberg2014speeding}, low-rank approximation through matrix factorization, such as singular-value decomposition, is used to factorize the weight matrices. 
The factors' rank is reduced by keeping only the leading eigenvalues and their associated eigenvectors. In \cite{compressvq}, quantization is used to reduce the storage taken by the model; both scalar quantization and vector quantization (VQ) have been considered. Using VQ, a weight matrix can be reconstructed from a list of indices and a dictionary of vectors. 
%While network compression methods do not generally accelerate inference, some of them do, or can be adapted to do so. For example, when using VQ as in \cite{compressvq}, one can avoid reconstructing the whole matrix with duplicate rows/columns, by using the codebook as the weight matrix and by altering the connectivity of the next layer. 
Thus, practical computation savings can be obtained.  Unfortunately, most network compression methods do not decrease the memory and compute usage during inference.  %In this work, we focus on methods that have the primary goal of accelerating inference and reducing memory usage.

\textbf{Neural network pruning} consists of identifying and removing neurons that are not necessary for achieving high performance. Some  of the first approaches used the second-order derivative to determine the sensitivity of the network to the value of each weight~\cite{lecun1990optimal,hassibi1993second}. A more recent, very simple and effective approach selects which neurons to remove by thresholding the magnitude of their weights; smaller magnitudes are associated with unimportant neurons~\cite{han2015learning}.  The resulting network is then finetuned for better performance. Nonetheless, experimental results (c.f. \secref{sec:experiments}) show that variational pruning methods (discussed below) outperform the previously mentioned works. %TODO verify claim about variational methods vs MorphNet

\textbf{Sparsity Learning (SL)} methods aim at pruning a network while training it.  Some methods add to the training loss a regularization function such as $L_1$~\cite{Liu2017}, Group LASSO~\cite{Wen16}, or an approximation of the $L_0$ norm~\cite{louizos2018learning, Pan16}.  Several variational methods have also been proposed~\cite{dai2018infobottlen, kingma2015variational,neklyudov2017structured, molchanov2017variational}.  These methods formalize the problem of network pruning as a problem of learning the parameters of a dropout probability density function (PDF) via the reparametrization trick~\cite{kingma2015variational}.  Pruning is enforced via a sparsity prior that derives from a variational evidence lower bound (ELBO).  %Other probability distributions have been explored, such as the Log-Normal distribution~\cite{neklyudov2017structured}, or a variation of the Binary Concrete distribution~\cite{maddison2016concrete, louizos2018learning}.
% TODO1: ou sont les bibtex manquants?
% TODO1: 'dropout function': premiere fois que je vois ce terme. Je crois qu'on parle ici d'appendre les paramètres d'une distribution
In general, SL methods do not apply an explicit constraint to limit the number of neurons used.  To enforce a budget, one has to turn towards budgeted pruning.%, which we discuss below.
% TODO better describe "sparsity prior" ?

% TODO
\textbf{Budgeted pruning} is an approach that provides a direct control on the size of the pruned network via some ``network size" hyper-parameter. %. Instead of using a \textit{speed-accuracy trade-off} hyper-parameter, like most pruning methods do, budgeted pruning offers a "network size" hyper-parameter. 
MorphNet \cite{gordon2018morphnet} alternates between training with a $L_1$ sparsifying regularizer and applying a width multiplier to the layer widths to enforce the budget. Contrary to our method, this work does not leverage dropout-based SL. Budgeted Super Networks \cite{veniat2018supernetworks} is a method that finds an architecture satisfying a resource budget by sparsifying a super network at the module level. This method is less convenient to use than ours, as it requires ``neural fabric" training through reinforcement learning. %In contrast, our method is compatible with all common architectures and regular supervised learning. 
Another budgeted pruning approach is ``Learning-Compression" \cite{carreira2018learning}, which uses the method of auxiliary coordinates~\cite{carreira2014distributed} instead of back-propagation. Contrary to this method, our approach adopts a usual gradient descent optimization scheme, and does not rely on the magnitude of the weights as a surrogate of their importance.

%\textbf{Architecture design} relies on using human intuition and trial and error to build more efficient neural networks. One aspect of network design is the size of the receptive fields. Nowadays, \nxn{3} filters are generally used at all depths of CNN models (aside from the first layer). While early works such as \cite{lecun1998gradient} used \nxn{5} filters inside the architecture, it is now generally admitted that a series of \nxn{3} convolutions achieves similar results with lower computational demands. Even smaller filters, such as such as \nxm{3}{1} or \nxn{1}, are used in modern models \cite{szegedy2015going,szegedy2016rethinking,szegedy2017inception}. Bottleneck layers \cite{resnet} replace a \nxn{3} convolution by a sequence of three convolutions: the first one is \nxn{1} and reduces the number of features, the next is \nxn{3}, and the last is \nxn{1} and recovers the initial number of features. This pattern is considered to provide the benefits of greater depth, while requiring similar compute. Iandola \etal~\cite{iandola2016squeezenet} explore many different approaches in the same direction. The pruning methods that we describe in this work can be applied to the previously mentioned architectures to push their efficiency even further.

\textbf{Architecture search} (AS) is an approach that led to efficient neural networks in terms of performance and parameterization. Using reinforcement learning and vast amounts of processing power, NAS \cite{nas_zoph2016neural} have learned novel architectures; some that advanced the state-of-the-art, others that had relatively few parameters compared to similarly effective hand-crafted models. PNAS \cite{nas_liu2017progressive} and ENAS \cite{nas_pham2018efficient} have extended this work by cutting the necessary compute resources. These works have been aggregated by EPNAS \cite{nas_perez2018efficient}. AS is orthogonal to our line of work as the learned architectures could be pruned by our method. In addition, AS is more complicated to implement as it requires learning a controller model by reinforcement learning. In contrast, our method features tools widely used in CNN training.

%,nas_liu2017progressive,nas_pham2018efficient,nas_perez2018efficient}.

\section{Our Approach} \label{ours}
\subsection{Dropout Sparsity Learning}

Before we introduce the specifics of our approach, let us first summarize the fundamental concepts of Dropout Sparsity Learning (DSL). 

Let $\mathbf{h}_l$ be the output of the $l$-th hidden layer of a CNN computed by $f_l(\mathbf{h}_{l-1})$, a transformation of the previous layer, typically a convolution followed by a batch norm and a non-linearity.  As mentioned before, one way of reducing the size of a network is by shutting down neurons with an element-wise product  $\odot$ between the output of layer $\mathbf{h}_{l-1}$ and a binary tensor $\mathbf{z}_{l-1}$:
\begin{eqnarray}
\mathbf{h}_l = f_l(\mathbf{h}_{l-1}\odot  \mathbf{z}_{l-1}).
\label{eq:dropout}
\end{eqnarray}

To enforce structured pruning and shutdown feature maps (not just individual neurons), % that shutting down individual neurons only reduces the model size and do not affect significantly the network computing power (FLOPs) and runtime memory footprint.  As such, one 
 one can redefine $\mathbf{z}_{l-1}$ as a vector of size $d_{l-1}$ where $d_{l-1}$ is the number of feature maps in $\mathbf{h}_{l-1}$. Then, $\mathbf{z}_{l-1}$ is applied over the spatial dimensions by performing an element-wise product with $\mathbf{h}_{l-1}$.

As one might notice, \eqref{eq:dropout} is the same as that of dropout~\cite{Srivastava14} for which  $\mathbf{z}_{l-1}$ is a tensor of independent random variables {\em i.i.d.} of a Bernoulli distribution $q(z)$.  To prune a network, DSL redefines $\mathbf{z}_{l-1}$ as random variables sampled from a distribution $q(z|\Phi)$ whose parameters $\Phi$ can be learned while training the model.  In this way, the network can learn which feature maps to drop and which ones to keep.%  For example, Dai et al.~\cite{dai2018infobottlen} use a Gaussian distribution ${\cal N}(z|\mathbf{\mu_{l-1,j}},\sigma_{l-1,j})$ whereas Louizos {\em et al}~\cite{louizos2018learning}, use a hard concrete distribution $HC(z|\alpha_{l-1,j})$, a smooth version of the Bernoulli distribution. 
 
Since the operation of sampling $\mathbf{z}_{l-1}$ from a distribution is not differentiable, it is common practice to redefine it with the reparametrization trick~\cite{kingma2015variational}:
\begin{eqnarray}
\mathbf{h}_l = f_l(\mathbf{h}_{l-1}\odot  g(\Phi_{l-1},\epsilon))
\label{eq:dropout2}
\end{eqnarray}
where $g$ is a continuous function differentiable with respect to $\Phi$ and stochastic with respect to $\epsilon$, a random variable typically sampled from $\mathcal{N}(0,1)$ or $\mathcal{U}(0,1)$.

In order to enforce network pruning, one usually incorporates a two-term loss :
\begin{eqnarray}
L(W,\Phi) = L_{D}(W,\Phi) + \lambda L_{S}(\Phi)
\end{eqnarray}
where $\lambda$ is the prior's weight, $W$ are the parameters of the network, $L_{D}(W,\Phi)$ is a {\em data loss} that measures how well the model fits the training data ({\em e.g.} the cross-entropy loss) and $L_{S}$ is a {\em sparsity loss} that measures how sparse the model is. While $L_S$ varies from one method to another, the KL divergence between $q(z|\Phi)$ and some prior distribution is typically used by variational approaches \cite{kingma2015variational,molchanov2017variational}.
Note that during inference, one can make the network deterministic by replacing the random variable $\epsilon$ by its mean.

\subsection{Soft and hard pruning}

As mentioned before, $g(\Phi_{l-1},\epsilon)$ is a continuous function differentiable with respect to $\Phi_{l-1}$.  Thus, instead of being binary, the pruning of \eqref{eq:dropout2} becomes continuous (soft pruning), so there is always a non-zero probability that a feature map will be activated during training.  
%The training process is divided into two stages. In the first stage, called "train-prune", feature maps are gated using a probability distribution. Thus, instead of being binary, the notion of pruning becomes continuous (soft pruning), and there is always a non-zero probability that a feature map will be "on" during this phase. 
However, to achieve practical speedups, one eventually needs to ``hard-prune" filters.  To do so, once training is over, the values of $\Phi$ are thresholded to select which filters to remove completely. % This threshold is set to the largest value of $\alpha$ for which $z_{\mathrm{infer}}(\alpha) = 0$ (see Appendix \ref{appen:z_inference}).During the second phase, called "fine-tune", the gates are binarized (hard pruning), the pruning objective is ignored, and training continues.
Then, the network may be fine-tuned for several epochs with the $L_{D}$ loss only, to let the network adapt to hard-pruning. We call this the ``fine-tuning phase", and the earlier epochs constitute the ``training phase".

\subsection{Budget-Aware Regularization (BAR)} \label{bar}

%Our Budget-Aware Regularization (BAR) method integrates a budget constraint.  
In our implementation, a budget is the maximum number of neurons a ``hard-pruned" network is allowed to have.  To compute this metric, one may replace $z \sim q(z|\Phi)$ by its mean so feature maps with $\mathbb{E}[z|\Phi] = 0$ have no effect and can be removed, while the others are kept. The network size is thus the total activation volume of the structurally ``hard-pruned" network : %footnote{Our choice of distribution $q(z|\Phi)$ \cite{louizos2018learning} can put significant weight on $z=0$; it can even put more than on all other values $z \in (0,1]$.}
\begin{equation} \label{eq:omega}
    V = \sum_l \sum_i \mathds{1}(\mathbb{E}[z_{l,i}|\Phi] > 0) \times A_l
\end{equation}
where $A_l$ is the area of the output feature maps of layer $l$ and $\mathds{1}$ is the indicator function. Our training process is described in Algorithm \ref{algo}.

\begin{algorithm}[t]
\DontPrintSemicolon
\SetKwData{TeacherLogits}{TeacherLogits}
\SetKwData{Prog}{Prog}
\SetKwData{Minibatches}{Minibatches}
\SetKwData{PrunedNet}{PrunedNet}
\SetKwData{PruningMasks}{PruningMasks}
\SetKwFunction{TrainUnprunedNetwork}{TrainUnprunedNetwork}
\SetKwFunction{PredictWholeDataset}{PredictWholeDataset}
\SetKwFunction{ForwardPass}{ForwardPass}
\SetKwFunction{BARLoss}{BARLoss}
\SetKwFunction{BackwardPass}{BackwardPass}
\SetKwFunction{OptimizationStep}{OptimizationStep}
\SetKwFunction{ConvertNet}{ConvertNet} % TODO Décrire?
\KwData{\footnotesize $W$: network weights; $\Phi$: r.v. parametrization; TeacherLogits: the class-wise scores for all samples of the dataset; $\boldsymbol{\lambda}$: all the hyperparameters of the method (including the budget); \Prog $\in [0,1]$:  progress of the training process; $g(\cdot)$: function introduced in Eq. (2) of the paper; $\mathbf{\hat{y}}$: predicted class-wise logits.}
\KwResult{\footnotesize PrunedNet: the pruned neural network object including its weights.}
\BlankLine

%$W' \Leftarrow$ \TrainUnprunedNetwork{}\;
$W' \Leftarrow$ TrainUnprunedNetwork()\;
%\TeacherLogits $\Leftarrow$ \PredictWholeDataset{$W'$}
TeacherLogits $\Leftarrow$ PredictWholeDataset($W'$)\;
\For{$b \in \mbox{Minibatches}$}{
	$(\mathbf{x},\mathbf{y}) \Leftarrow b$\;
	$\mathbf{z} \Leftarrow g(\Phi, \boldsymbol{\epsilon}), \boldsymbol{\epsilon} \sim \mathcal{U}(0,1)$\;
%	$\mathbf{\hat{y}} \Leftarrow$ \ForwardPass{$\mathbf{x}, W, \mathbf{z}$}\;
	$\mathbf{\hat{y}} \Leftarrow$ ForwardPass($\mathbf{x}, W, \mathbf{z}$)\;
%	$l \Leftarrow$ \BARLoss{$\mathbf{y}, \mathbf{\hat{y}}, \mathbf{z}, \boldsymbol{\lambda}, \TeacherLogits, \Prog$}\;
    $l \Leftarrow$ BARLoss($\mathbf{y}, \mathbf{\hat{y}}, \mathbf{z}, \boldsymbol{\lambda}, \mbox{TeacherLogits}, \Prog$)\;
%	$(\nabla W, \nabla \Phi) \Leftarrow$ \BackwardPass{$l$}\;
    $(\nabla W, \nabla \Phi) \Leftarrow$ BackwardPass($l$)\;
%	$(W, \Phi) \Leftarrow$ \OptimizationStep{$\nabla W, \nabla \Phi$}\;
    $(W, \Phi) \Leftarrow$ OptimizationStep($\nabla W, \nabla \Phi$)\;
}
PruningMasks $\Leftarrow g(\Phi, \mathbb{E}[\epsilon])$\;
PrunedNet $\Leftarrow$ ConvertNet($W$,\ PruningMasks)\;
\caption{BAR Training}\label{algo}\end{algorithm}

A budget constraint imposes on $V$ to be smaller than the allowed budget $b$.  If embedded in a sparsity loss, that constraint makes the loss go to infinity when $V>b$, and zero otherwise.  This is a typical inequality constrained minimization problem whose binary (and yet non-differentiable) behavior is not suited to gradient descent optimization. One typical solution to such problem is the {\em log-barrier} method~\cite{Boyd2004}. The idea of this barrier method is to approximate the zero-to-infinity constraint by a differentiable logarithmic function : $-(1/t)\log(b-V)$ where $t > 0$ is a parameter that adjusts the accuracy of the approximation and whose value increases at each optimization iteration (c.f. Algo 11.1 in~\cite{Boyd2004}).

Unfortunately, the log-barrier method requires beginning optimization with a feasible solution (i.e. $V < b$), and this brings two major problems. First, we need to compute $\Phi$ such that $V < b$, which is no trivial task. Second, this induces a setting similar to training an ensemble of pruned networks, as the probability that a feature map is ``turned on" is very low. This means that filters will receive little gradient and will train very slowly. To avoid this, we need to start training with a $V$ larger than the budget.

\begin{figure}[t]
\vspace{-1em}
  \centering
  \subfloat[Logarithmic barrier function]{\includegraphics[width=0.49\linewidth]{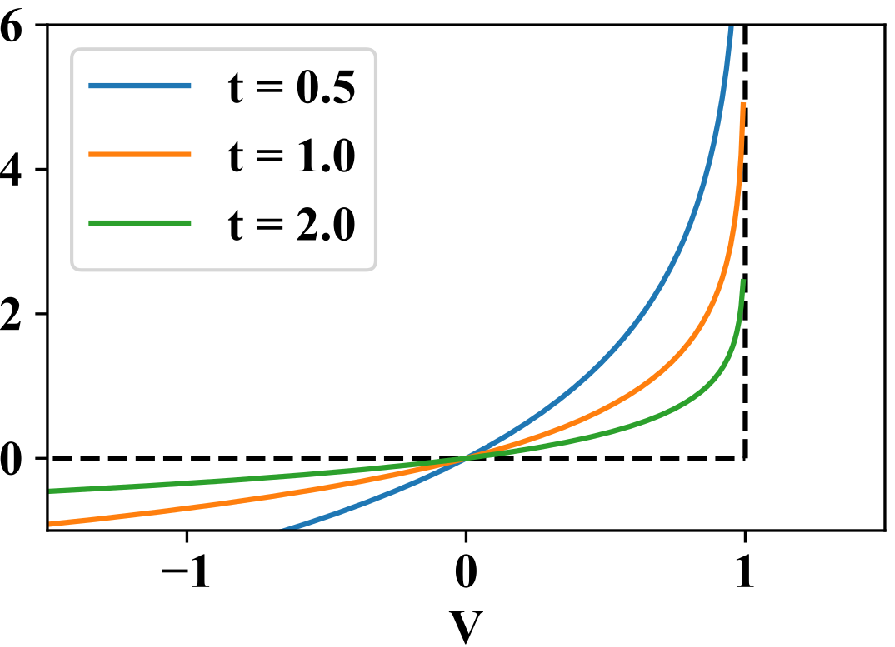}}
  \hspace{2pt}
  \subfloat[Our barrier function]{\includegraphics[width=0.49\linewidth]{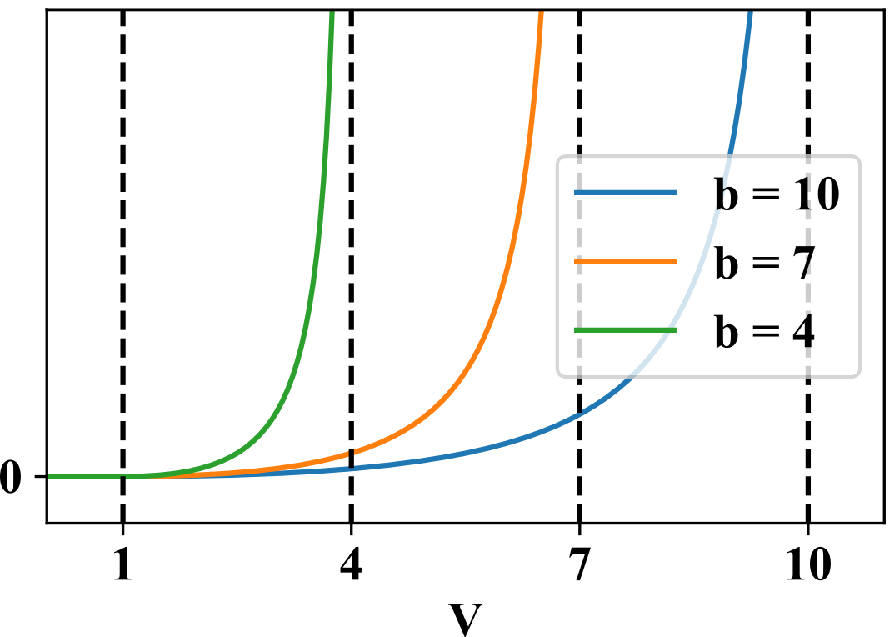}}
   \caption{\textbf{Comparing barrier functions}. \textit{(a)} Common barrier function $-(1/t)\log(b-V)$ with $b=1$. \textit{(b)} Our barrier function $f(V, a, b)$ with $a=1$. \vspace{-0.4cm}}%. Notice the asymptote at $V=b$.\vspace{-0.4cm}}
   \label{fig:barrier}
\end{figure}

We thus implemented a modified version of the barrier algorithm.  First, as will be shown in the rest of this section, we propose a barrier function  $f(V, a, b)$ as a replacement for the log barrier function (c.f. \figref{fig:barrier}). Second, instead of having a fixed budget $b$ and a parameter $t$ that grows at each iteration as required by the barrier method, we eliminate the hardness parameter $t$ and instead decrease the budget constraint at each iteration. This budget updating schedule is discussed in \secref{pruningtargettrans}.

Our barrier function $f(V, a, b)$ is designed such that: \vspace{-0.3cm}
\begin{itemize}
\item it has an infinite value when the volume used by a network exceeds the budget, {\em i.e. $V>b$}; \vspace{-0.25cm}
\item it has a value of zero when the budget is \textit{comfortably} respected, {\em i.e. $V<a$}; \vspace{-0.25cm}
\item it has $C^1$ continuity.
\end{itemize}
%

% \begin{figure}
% \begin{center}
% 	 \includegraphics[width=\linewidth]{fVab.svg.eps}
% 	 \vspace{-2em}
% \end{center}
%   \caption{\textbf{An illustrative plot of $f(V,a,b)$}. This function is used to smooth the budget constraint. Notice the asymptote at $V=b$. Recall that $f(V, a, b) = \infty \ \forall \ V \geq b$.}
%   \label{fig:barrier}
% \end{figure}

Instead of having a jump from zero to infinity at the point where $V>b$, we define a range where a smooth transition occurs.  To do so, we first perform a linear mapping of $V$~:
\begin{equation*}
    c = \frac{V - a}{b - a}
\end{equation*}
such that $V = a \Rightarrow c=0$ (the budget is \textit{comfortably} respected), and $V = b \Rightarrow c=1$ (our constraint $V < b$ is violated). Then, we use the following function:
\begin{equation*}
    g(c) = \frac{c^2}{1-c}
\end{equation*}
which has three useful properties: ($i$) $g(0)=0$ and $g(0)'=0$, ($ii$) $\lim_{c \to 1^{-}} g(c) = \infty$ and ($iii$) it has a $C^1$ continuity. Those properties correspond to the ones mentioned before. To obtain the desired function, we substitute $c$ in $g(c)$ and simplify:
\begin{equation} 
f(V,a,b) =  \begin{cases} 
      0 & V \leq a \\
      \frac{(V-a)^2}{(b-V)(b-a)} & a < V < b \\
      \infty & V \geq b.
      \end{cases}
\end{equation}

%\begin{equation*}
%\begin{split}
%    f(V,a,b) &= \frac{(\frac{V - a}{b - a})^2}{1-(\frac{V - a}{b - a})} = \frac{(\frac{V - a}{b - a})^2}{\frac{b - a - (V - a)}{b - a}} \\
%    &= \frac{(V-a)^2(b-a)}{(b-a)^2(b-V)}
%    = \frac{(V - a)^2}{(b-V)(b - a)}.
%\end{split}
%\end{equation*}

As shown in \figref{fig:barrier}, like for log barrier, $V=b$ is an asymptote, as we require $V < b$. However, $a<V < b$ corresponds to a respected budget and for $V\leq a$, the budget is respected with a comfortable margin, and this corresponds to a penalty of zero.%  Note that in practice, we clamp $f(V, a, b)$ to a very large value that does not overflow the numerical representation, such as $10^{10}$. %Note that this formulation requires to have a feasible initial solution. We will explain in \secref{pruningtargettrans} how this is achieved.

Our proposed prior loss is as follows:
\begin{equation} \label{eq:objfn}
%\begin{split}
L_{\text{BAR}}(\Phi,V,a,b) = L_{S}(\Phi)f(V,a,b) %\\
%f(V,a,b) &=  \begin{cases} 
%      0 & V \leq a \\
%      \frac{(V-a)^2}{(b-V)(b-a)} & a < V < b \\
%      \infty & V \geq b
%   \end{cases}
%\end{split}
\end{equation}

where $(a,b)$ are the lower and upper budget margins, $V$ is the current ``hard-pruned" volume as computed by \eqref{eq:omega}, and $L_{S}(\Phi)$ is a differentiable approximation of $V$. Note that since $V$ is not differentiable w.r.t to $\Phi$, we cannot solely optimize $f(V,a,b)$.

The content of $L_{S}(\Phi)$ is bound to $q(z|\Phi)$.  In our case, we use the Hard-Concrete distribution (which is a smoothed version of the Bernoulli distribution), as well as its corresponding prior loss, both introduced in \cite{louizos2018learning}. This prior loss measures the expectation of the number of feature maps currently unpruned. To account for the spatial dimensions of the output tensors of convolutions, we use:
\begin{equation*}
    L_{S}(\Phi) = \sum_l L_{S}(\Phi_l) = \sum_l L_{HC}(\Phi_l) \times A_l
\end{equation*}
where $L_{HC}$ is the hard-concrete prior loss~\cite{louizos2018learning} and $A_l$ is the area of the output feature maps of layer~$l$. Thus, $L_{S}(\Phi)$ measures the expectation of the activation volume of all convolution operations in the network.

Note that $V$ could also be replaced by another metric, such as the total FLOPs used by the network. In this case,  $L_{S}(\Phi_l)$ should also include the expectation of the number of feature maps of the preceding layer.% in order to approximate the FLOPS. %In this work, however, we optimize the total activation volume as a simple proxy to the total FLOPS.

%To obtain the ``hard-pruned" network, we replace $z \sim q(z|\Phi)$ by its mean; thus, feature maps with $\mathbb{E}[z|\Phi] = 0$ have no effect and can be removed, while the others are kept. The total activation volume of the ``hard-pruned" network is computed with the following:

%\begin{equation} \label{eq:omega}
%    V = \sum_l \sum_i \mathds{1}(\mathbb{E}[z|\Phi_{l,i}] > 0) \times A_l
%\end{equation}

\subsection{Setting the budget margins $(a,b)$} \label{pruningtargettrans}
As mentioned earlier, initializing the network with a volume that respects the budget (as required by the barrier method) leads to severe optimization issues.  Instead, we iteratively shift the pruning target $b$ during training. Specifically, we shift it from $b=V_F$ at the beginning, to $b=B$ at the end (where $V_F$ is the unpruned network's volume and $B$ the maximum allowed budget). 

As shown in \figref[b]{fig:barrier}, doing so induces a lateral shift to the ``barrier". This is unlike the barrier method in which the hardness parameter $t$ evolves in time (c.f. \figref[a]{fig:barrier}). Mathematically, the budget $b$ evolves as follows:
\begin{equation} \label{eq:pruningtarget}
\begin{split}
%a &=B-0.0001\ V_F,\\
b &= (1-T(i))\ V_F + T(i)\ B, \\
i &= \frac{\mathrm{iteration\ index}}{\mathrm{num.\ training\ iterations}}
\end{split}
\end{equation}
while $a=B-10^{-4}\ V_F$ is fixed.  Here $T(i)$ is a \textit{transition function} which goes from zero at the first iteration all the way to one at the last iteration. While $T(i)$ could be a linear transition schedule, experimental results reveal that when $b$ approaches $B$, some gradients suffers from extreme spikes due to the nature of $f(V,a,b)$.  This leads to erratic behavior towards the end of the training phase that can hurt performance.
%In an attempt to alleviate this excessive stress, one 
One may also implement an exponential transition schedule. This could compensate for the shape of $f(V,a,b)$ by having $b$ change quickly during the first epochs and slowly towards the end of training.  While this gives good results for severe pruning (up to $16\times$), the increased stress at the beginning yields sub-optimal performance for low pruning factors.

\begin{figure}[t]
\begin{center}
	 \vspace{-1em}
	 \includegraphics[width=0.7\linewidth]{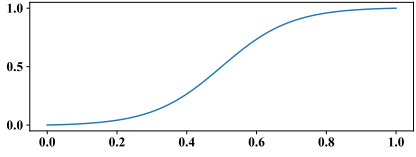}
	 \vspace{-0.4cm}
\end{center}
   \caption{\textbf{Sigmoidal transition function.}\vspace{-0.4cm}}
   \label{fig:sigmoid}
\end{figure}

For our method, we propose a sigmoidal schedule, where $b$ changes slowly at the beginning and at the end of the training phase, but quickly in the middle. This puts most of the ``pruning stress" in the middle of the training phase, which accounts for the difficulty of pruning (1) during the first epochs, where the filters' relevance is still unknown, and (2) during the last epochs, where more compromises might have to be made. The sigmoidal transition function is illustrated in \figref{fig:sigmoid} (c.f. Supp. materials for details).

\subsection{Knowledge Distillation} \label{kds}

% refer reader to original (or other) article for introduction?
Knowledge Distillation (KD) \cite{hinton2015distilling} is a method for facilitating the training of a small neural network (the \textit{student}) by having it reproduce the output of a larger network (the \textit{teacher}). %More specifically, it is used to train a student network that is smaller than its teacher; hence the term \textit{distillation}. 
%While normal classification training minimizes the cross entropy between the prediction vector and the target \textit{one-hot} vector, 
%To do so, KD minimizes the divergence between the class-wise predictions of the student and the teacher.  
%The efficiency of this approach at training smaller networks could be explained by the following two reasons. Firstly, all neurons of the classification layer and their respective paths in the network are trained for every training sample. Without KD, only the neuron associated with the correct class has non-zero loss, and thus back-propagates a loss signal. %TODO a vérifier
%Secondly, the predictions of the teacher are helpful in cases where two classes are hard to distinguish. For example, a good image classifier is likely to assign a substantial score to the class \textit{car} when it is given the image of a \textit{pick-up}. Mixing up these two classes not a big mistake at first when learning to distinguish between many object classes; but without KD, mixing up \textit{car} and \textit{pick-up} is as big of a mistake as mixing up \textit{car} and \textit{bird}.
The loss proposed by Hinton et al~\cite{hinton2015distilling} is :
\begin{eqnarray}
L_{D}(W) = (1-\alpha)L_{CE}(P_s,Y_{gt}) + \alpha T^2 L_{CE}(P_s,P_t) \nonumber
\end{eqnarray}
where $L_{CE}$ is a cross-entropy, $Y_{gt}$ is the groundtruth, $P_s$ and $P_t$ are the output logits of the student and teacher networks, $\alpha\in [0,1]$, and $T \geq 1$ is a temperature parameter used to smooth the softmax output : $p_i = \frac{\exp(z_i/T)}{\sum_j\exp(z_j/T)}$.

In our case, the unpruned network is the \textit{teacher} and the pruned network is the \textit{student}.  As such, our final loss is:
\begin{equation*}
   (1-\alpha)L_{CE}(P_s,Y_{gt}) + \alpha T^2 L_{CE}(P_s,P_t)  + \lambda L_{\text{BAR}}(\Phi,V,a,b). 
\end{equation*}
where $\lambda$, $\alpha$ and $T$ are fixed parameters.
%
%where $\lambda = 10^{-5}$,  $\alpha = 0.9$, and $T = 4$ (found empirically).
%where the values of $\lambda$, $k_\alpha$ and $k_T$ have been found empirically.
%TODO put everything: KD, cross-entropy, scheduling, hard concrete distribution, etc.}

% \subsection{Inflating the pruned network}

% Our method always produce networks with fewer neurons than the allowed budget $B$, and the removal of ``orphan" convolutions (where the previous or next layers are completely pruned) further reduces the number of neurons used. Thus, we can ``revive" some filters  in a post-processing stage to fully use the budget. To do so, we search a threshold on $\Phi$ that allows the existence of the most filters while respecting the budget. %Searching a threshold on $\mathbb{E}[z|\Phi]$  would not help to recover the best of the ``dead" features, as $\mathbb{E}[z|\Phi] = 0$ for all pruned filters. 
% After finding the threshold on $\Phi$ through a binary search, we binarize the gates such that $z \in \{0,1\}$; this ensures all unpruned feature maps can output a strong signal. This post-processing operation takes roughly a second to compute.

\section{Pruning Residual Networks}

While our method can prune any CNN, pruning a CNN without residual connections does not affect the connectivity patterns of the architecture, and simply selects the width at each layer~\cite{gordon2018morphnet}. %TODO pourquoi citer morphnet ici?
In this paper, we are interested in allowing any feature map of a residual network to be pruned. This pruning regime can reduce the depth of the network, and generally results in architectures with atypical connectivity that require special care in their implementation to obtain maximum efficiency.

%\vspace{-0.1cm}
\subsection{Automatic Depth Determination} \label{depth_selection}
%\vspace{-0.2cm}

We found, as in \cite{gordon2018morphnet}, that filter-wise pruning can successfully prune entire ResBlocks and change the network depth. This effect was named \textit{Automatic Depth Determination} in~\cite{unifying_dropout}.  Since a ResBlock computes a delta that is aggregated with the main (residual) signal by addition (c.f. \figref[a]{fig:resblock}), such block can generally be removed without preventing the flow of signal through the network. This is because the main signal's identity connections cannot be pruned as they lack prunable filters.

\begin{figure}[t]
\begin{center}
	 \includegraphics[width=0.9\linewidth]{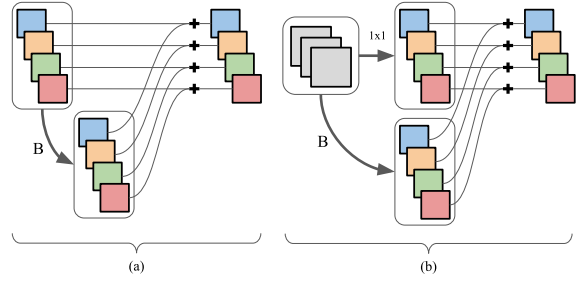}\vspace{-0.3cm}
\end{center}
   \caption{\textbf{Typical ResBlock vs. pooling block}. \textit{(a)} A typical ResBlock. The ``B" arrow is the sequence of convolutions done inside the block. \textit{(b)} A pooling block at the beginning of a ResNet Layer, that deals with the change in spatial dimensions and number of feature maps. Notice that it breaks the continuity of the residual signal. The arrow labeled ``\nxn{1}" is a \nxn{1} convolution with stride 2; the first convolution of ``B" also has stride 2. If all convolutions (arrows) are removed, no signal can pass.\vspace{-0.3cm} }
   \label{fig:resblock}
\end{figure}

However, some ResBlocks, which we call ``pooling blocks", change the spatial dimensions and feature dimensionality of the signal. This type of block breaks the continuity of the residual signal (c.f. \figref[b]{fig:resblock}). As such, the convolutions inside this block cannot be completely pruned, as this would prevent any signal from flowing through it (a situation we call ``fatal pruning").  As a solution, we clamp the highest value of $\Phi$ to ensure that at least one feature map is kept in the \nxn{1} conv operation.

\subsection{Atypical connectivity of pruned ResNets} \label{atypical}
\begin{figure}[t]
\begin{center}
	 \includegraphics[width=0.9\linewidth]{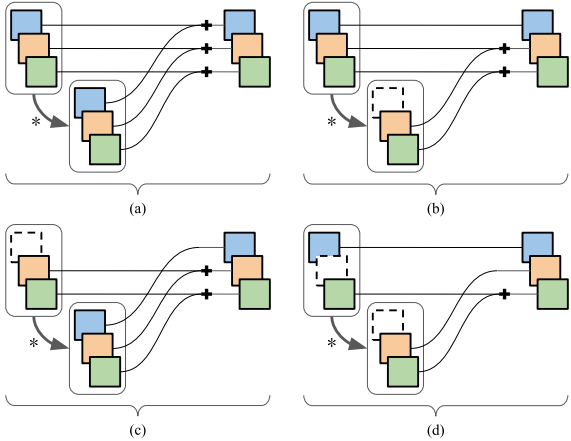}\vspace{-0.4cm}
\end{center}
   \caption{\textbf{Connectivity allowed by our approach}. \textit{(a)} A 3-feature ResBlock with typical connectivity. Arrows represent one or more convolutions. \textit{(b)} With one feature map pruned, only two features are computed and added to the residual signal; one feature from the residual signal is left unchanged. \textit{(c)} a new feature is created and concatenated to the residual signal. \textit{(d)} a combination of (b) and (c) as a new feature is concatenated to the residual signal, one feature from the residual is left unchanged, and a third feature has typical connectivity (best viewed in color).\vspace{-0.4cm}}
   \label{fig:exotic}
\end{figure}

%\vspace{-0.2cm}
Our method allows any feature map in the output of a convolution to be pruned (except for the \nxn{1} conv of the pooling block). This produces three types of atypical residual connectivity that requires special care (see \figref{fig:exotic}). For example, there could be a feature from the residual signal that would pass through without another signal being added to it (\figref[b]{fig:exotic}).  New feature maps can also be created and concatenated (\figref[c]{fig:exotic}).  Furthermore, new feature maps could be created while others could pass through (\figref[d]{fig:exotic}). 

To leverage the speedup incurred by a pruned feature map, the three cases in \figref{fig:exotic} must be taken into account through a mixed-connectivity block which allows these unorthodox configurations. Without this special implementation, some zeroed-out feature maps would still be computed because the summations of residual and refinement signals must have the same number of feature maps.  In fact, a naive implementation does not allow refining only a subset of the features of the main signal (as in \figref[b]{fig:exotic}), nor does it allow having a varying number of features in the main signal (as in \figref[c]{fig:exotic}). 

\begin{figure*}[t]
\begin{center}
	 \includegraphics[width=0.9\linewidth]{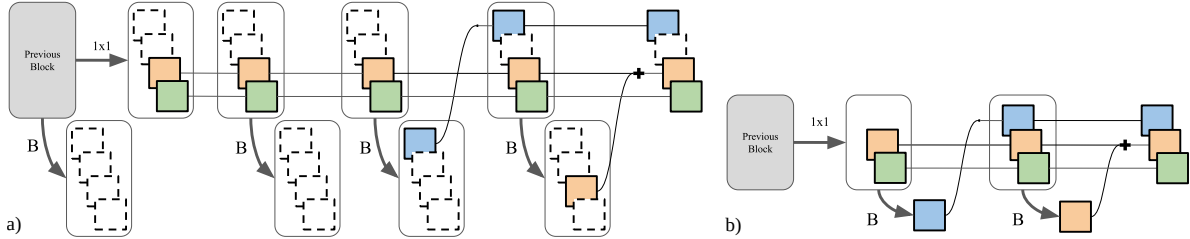} \vspace{-0.3cm}
\end{center}
   \caption{(a) \textbf{A 4-feature chunk of a ResNet Layer pruned by our method}. Dotted feature maps are zeroed-out by their associated mask. An arrow labeled B represents a Block operation, which consist of a sequence of convolutions. Inner convolutions of the Block can be pruned, but only the output of the last convolution is shown (for clarity). (b)~\textbf{The same pruned subgraph}, illustrated without the pruned feature maps. The resulting subgraph is shallower and narrower than its ``full" counterpart (best viewed in color). \vspace{-0.4cm} }
  \label{fig:example}
\end{figure*}

\figref{fig:example} shows the benefit of a mixed-connectivity block. In (a) is a ResNet Layer pruned by our method. Using a regular ResBlock implementation, all feature maps in pairs of tensors that are summed together need to have matching width. This means that, in \figref{fig:example}, all feature maps of the first, third and fourth rows (features) are computed, even if they are dotted. Only the second row can be fully removed.%; however, there are no restrictions on pruning in inner convolutions of ResBlocks. 
On the other hand, by using mixed-connectivity, only unpruned feature maps are computed, yielding architectures such as in \figref[b]{fig:example}, that saves substantial compute (c.f. \secref{sec:experiments}).

Technical details on our mixed-connectivity block are provided in the Supplementary materials.

\section{Experiments} \label{sec:experiments}
\subsection{Experimental Setup}
We tested our pruning framework on two residual architectures and report results on four datasets. We pruned Wide-ResNet~\cite{wrn} on CIFAR-10, CIFAR-100 and TinyImageNet (with a width multiplier of 12 as per \cite{wrn}), and ResNet50 \cite{resnet} on Mio-TCD~\cite{Luo18}, a larger and more complex dataset devoted to traffic analysis. TinyImageNet and Mio-TCD samples are resized to \nxn{64} and \nxn{128}, respectively. Since this ResNet50 has a larger input and is deeper than its CIFAR counterpart, we do not opt for the ``wide" version and thus save significant training time. Both network architectures have approximately the same volume.

For all experiments, we use the Adam optimizer with an initial learning rate of $10^{-3}$ and a weight decay of $5 \times 10^{-4}$. For CIFAR and TinyImageNet, we use a batch size of 64. For our objective function, we use $\alpha = 0.9$, $T = 4$, and $\lambda = 10^{-5}$. We use PyTorch and its standard image preprocessing. For experiments on Mio-TCD, we start training/pruning with the weights of the unpruned network whereas we initialize with random values for CIFAR and TinyImageNet. Please refer to the Supplementary materials for the number of epochs used in each training phase. %TODO référencer?
% We train for a total of 110 epochs on CIFAR-10 and CIFAR-100, and for a total of 70 epochs on Mio-TCD. We finetune for 40 for CIFAR-10 and CIFAR-100, and for 20 epochs for Mio-TCD. We divide the learning rate by ten for the last ten epochs for all datasets.

We compare our approach to the following methods:\vspace{-0.2cm}

\begin{itemize}
    \item \textbf{Random}. This approach randomly selects feature maps to be removed. \vspace{-0.2cm}
    \item \textbf{Weight Magnitude (WM)} \cite{han2015learning}. This method uses the absolute sum of the weights in a filter as a surrogate of its importance. Lower magnitude filters are removed.\vspace{-0.2cm}
    \item \textbf{Vector Quantization (VQ)} \cite{compressvq} This approach vectorizes the filters and quantizes them into $N$ clusters, where $N$ is the target width for the layer. The clusters' center are used as the new filters.\vspace{-0.2cm}
    \item \textbf{Interpolative Decomposition (ID)}. This method is based on low-rank approximation for network compression \cite{denton2014exploiting,jaderberg2014speeding}. This algorithm factorizes each filters $W$ into $UV$, where $U$ has a specific number of rows corresponding to the budget. $U$ replaces $W$, and $V$ is multiplied at the next layer (i.e. $W_{l+1} \gets V_l W_{l+1}$) to approximate the original sequence of transformations. \vspace{-0.4cm}
    \item \textbf{$L_0$ regularization (LZR)} \cite{louizos2018learning}. This DSL method is the closest to our method. However, it incorporates no budget, penalizes layer width instead of activation tensor volume, and does not use Knowledge Distillation.\vspace{-0.2cm}
    \item \textbf{Information Bottleneck (IB)} \cite{dai2018infobottlen}. This DSL method uses a factorized Gaussian distribution (with parameters $\mu,\sigma$) to mask the feature maps as well as the following prior loss :  $L_S=\log(1-\mu^2 / \sigma^2)$.\vspace{-0.2cm}% and suggest to set the hard pruning threshold by visual inspection of the empirical values. We make this process automatic by using an agglomerative clustering algorithm \footnote{The authors say that the "alive" and "dead" filters are well separated. We find these clusters using \textit{scikit-learn} ( http://scikit-learn.org/ stable/modules/generated/sklearn.cluster.AgglomerativeClustering.html)}. This method has an hyperparameter that requires a sweep.
    \item \textbf{MorphNet} \cite{gordon2018morphnet}. This approach uses the $\gamma$ scaling parameter of Batch Norm modules as a learnable mask over features. The said $\gamma$ parameters are driven to zero by a $L_1$ objective that considers the resources used by a filter (e.g. FLOPs). This method computes a new width for each layer by counting the non-zero $\gamma$ parameters. We set the sparsity trade-off parameter $\lambda$ after an hyperparameter search, with $16\times$ as the target pruning factor for CIFAR-10.\vspace{-0.2cm}
\end{itemize}

For every method, we set a budget of tensor activation volume corresponding to $1/2, 1/4, 1/8, 1/16$ of the unpruned volume~$V_F$.
Since {\em LZR} and {\em IB} do not allow setting a budget, we went through trial-and-error to find the hyperparameter value that yield the desired resource usage. For {\em Random, WM, VQ}, and {\em ID} we scale the width of all layers uniformly to satisfy the budget and implement a pruning scheme which revealed to be the most effective (c.f. Supplementary materials).  We also apply our mixed-connectivity block to the output of every method for a fair comparison. 

\subsection{Results}

\begin{figure*}[t]
\begin{center}
	 \includegraphics[width=0.248\linewidth]{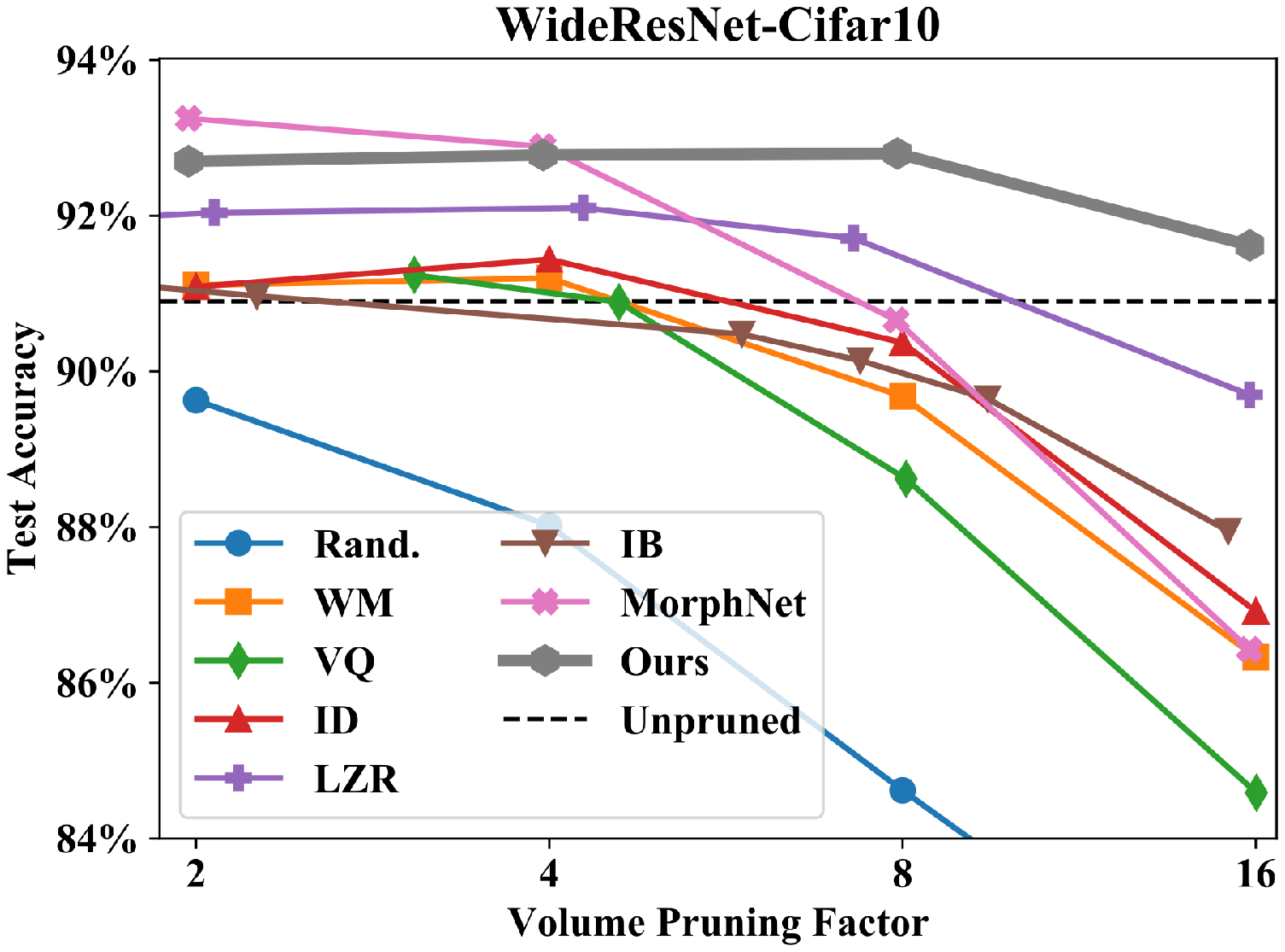}%\vspace{-0.3cm}
	 \includegraphics[width=0.248\linewidth]{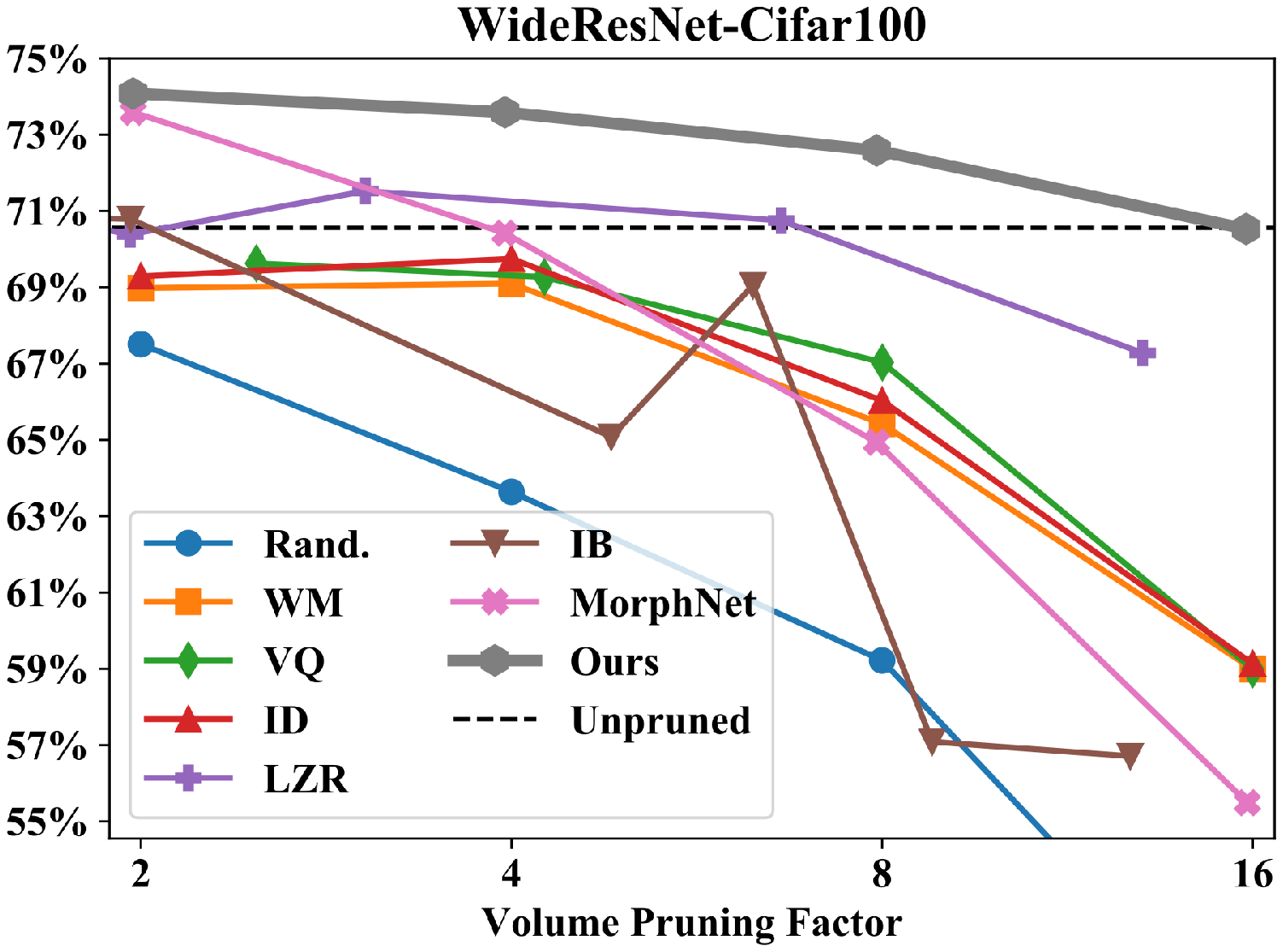}%\vspace{-0.3cm}
	 \includegraphics[width=0.248\linewidth]{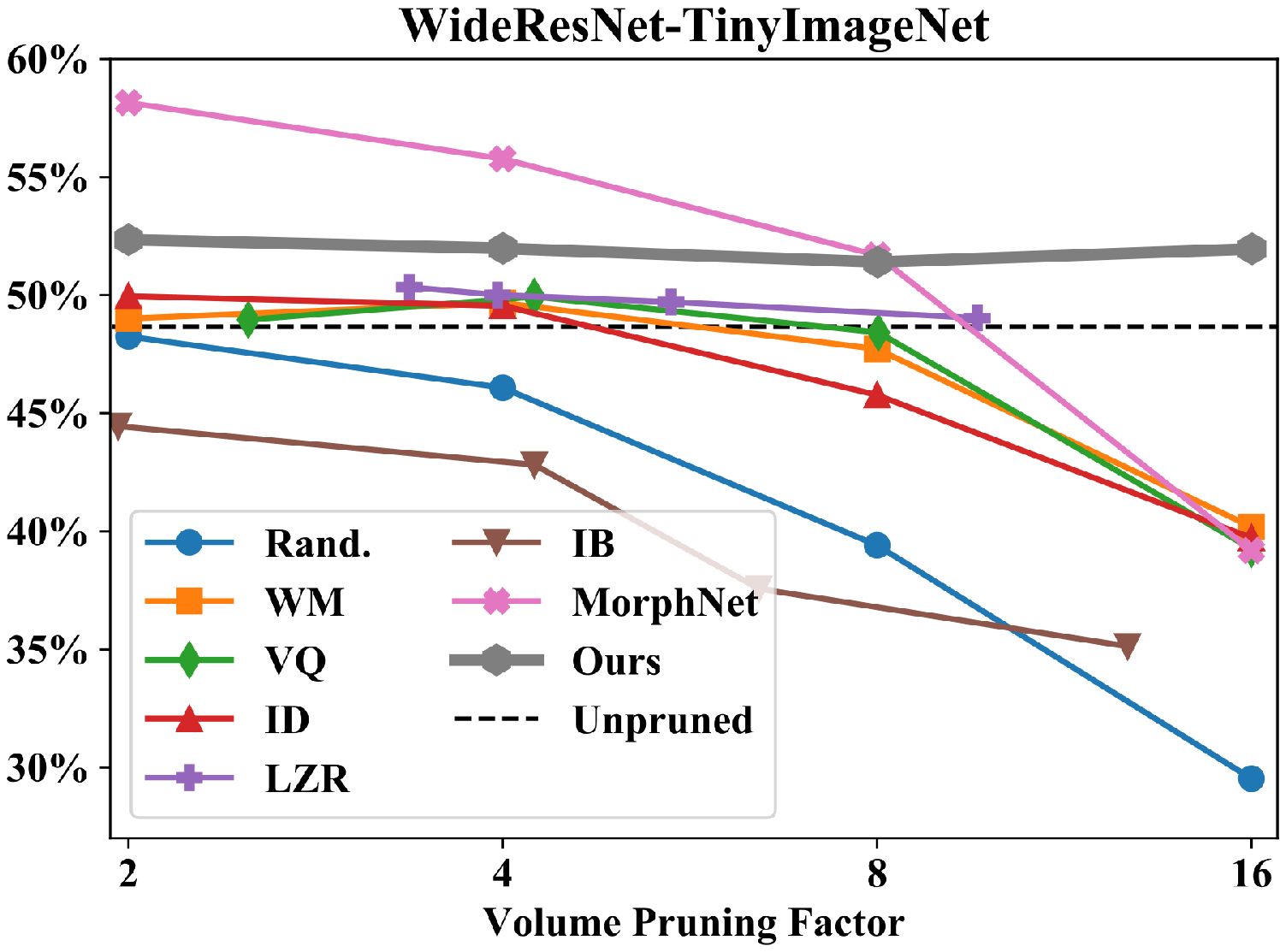}
	 \includegraphics[width=0.248\linewidth]{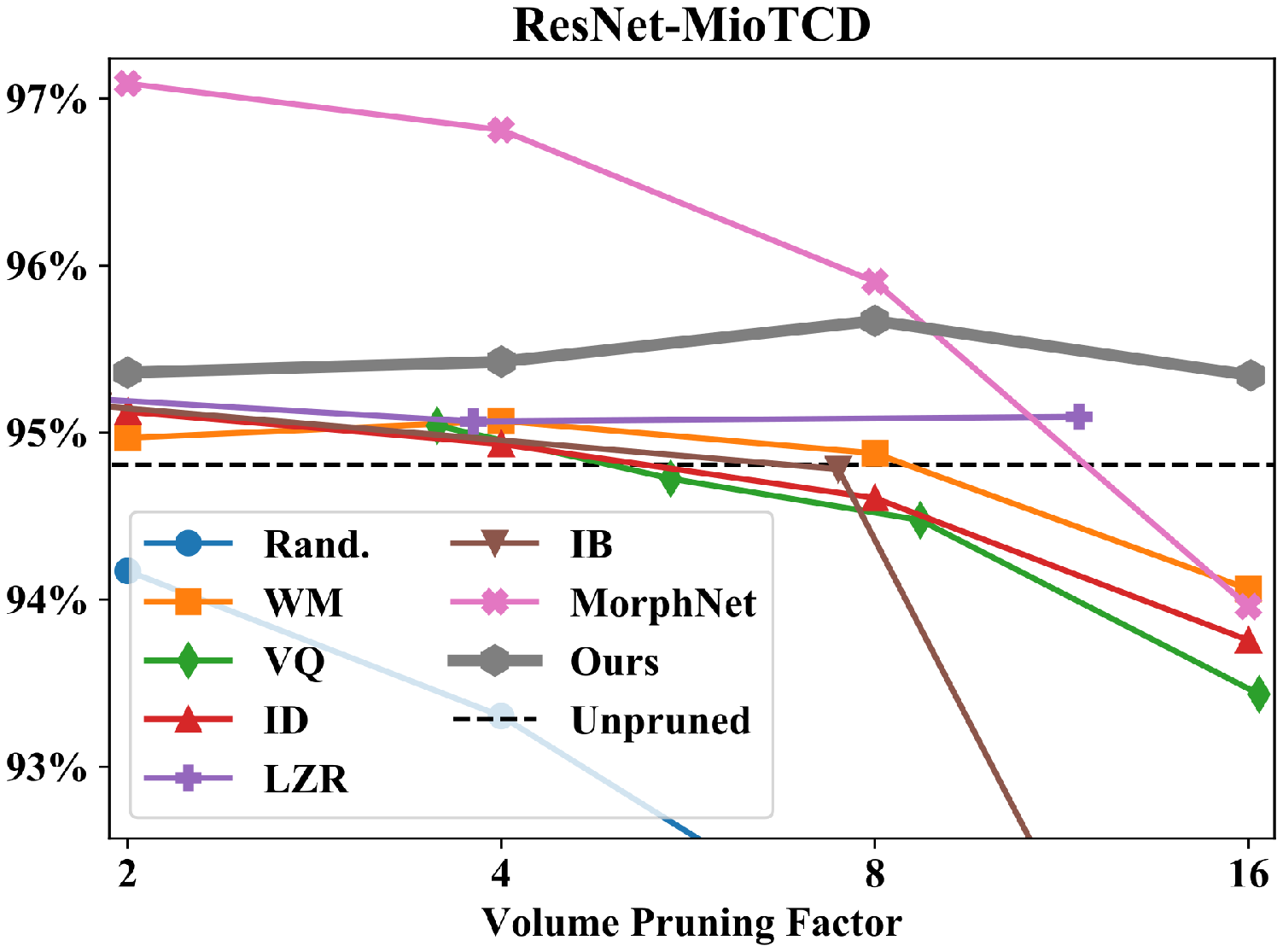}\\%\vspace{-0.5cm}
	 \includegraphics[width=0.248\linewidth]{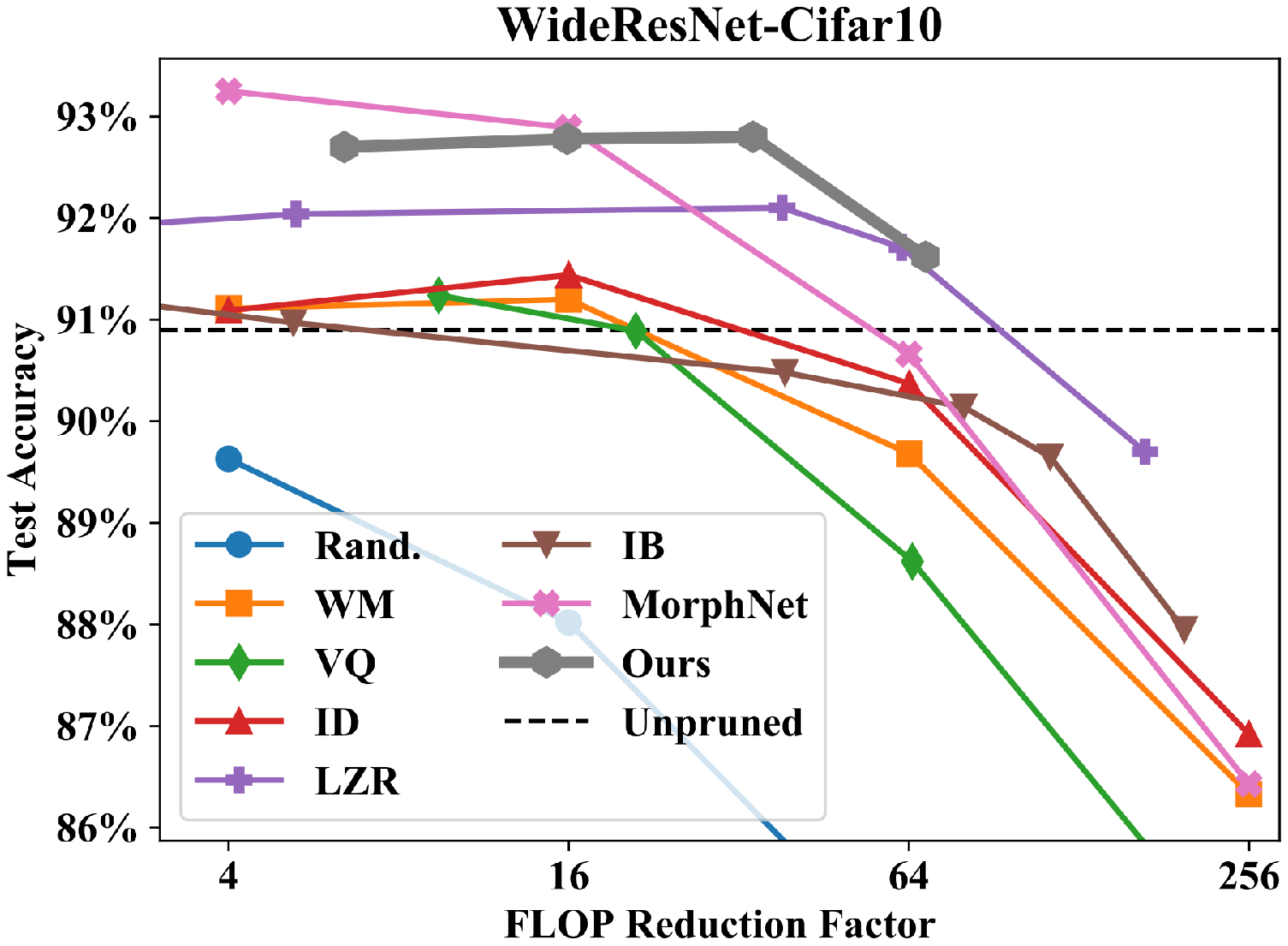}%\vspace{-0.3cm}
	 \includegraphics[width=0.248\linewidth]{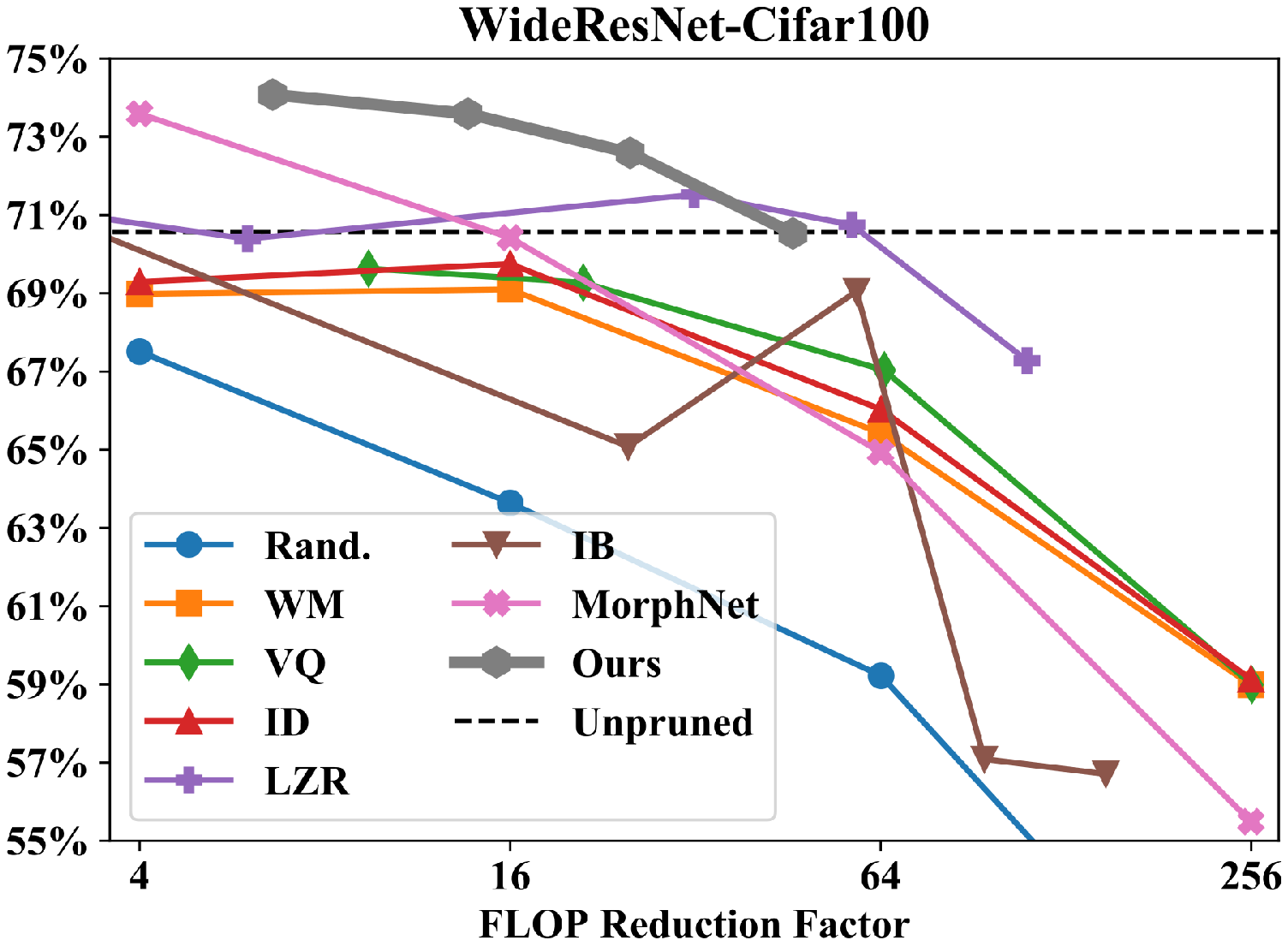}%\vspace{-0.3cm}
	 \includegraphics[width=0.248\linewidth]{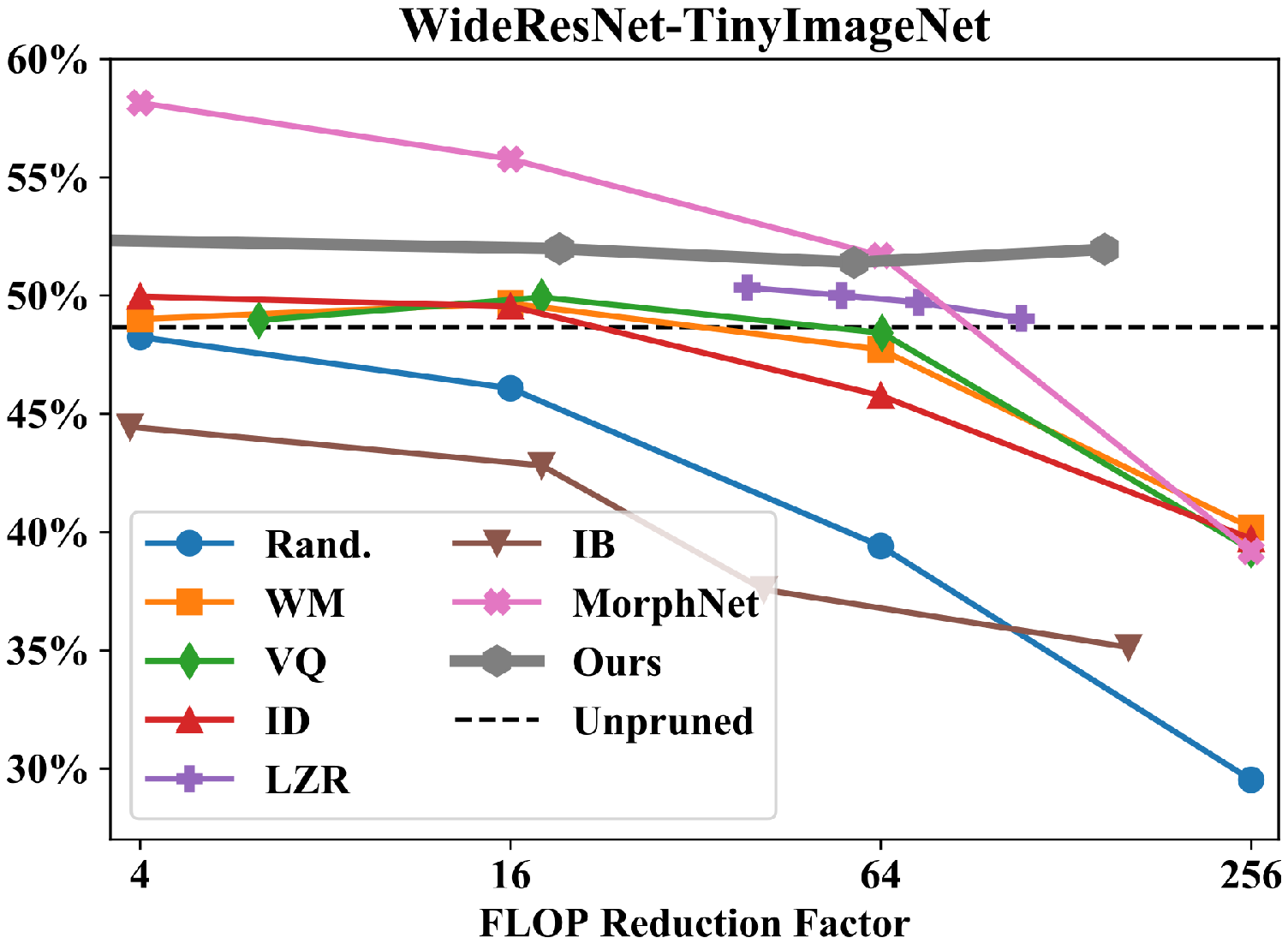}
	 \includegraphics[width=0.248\linewidth]{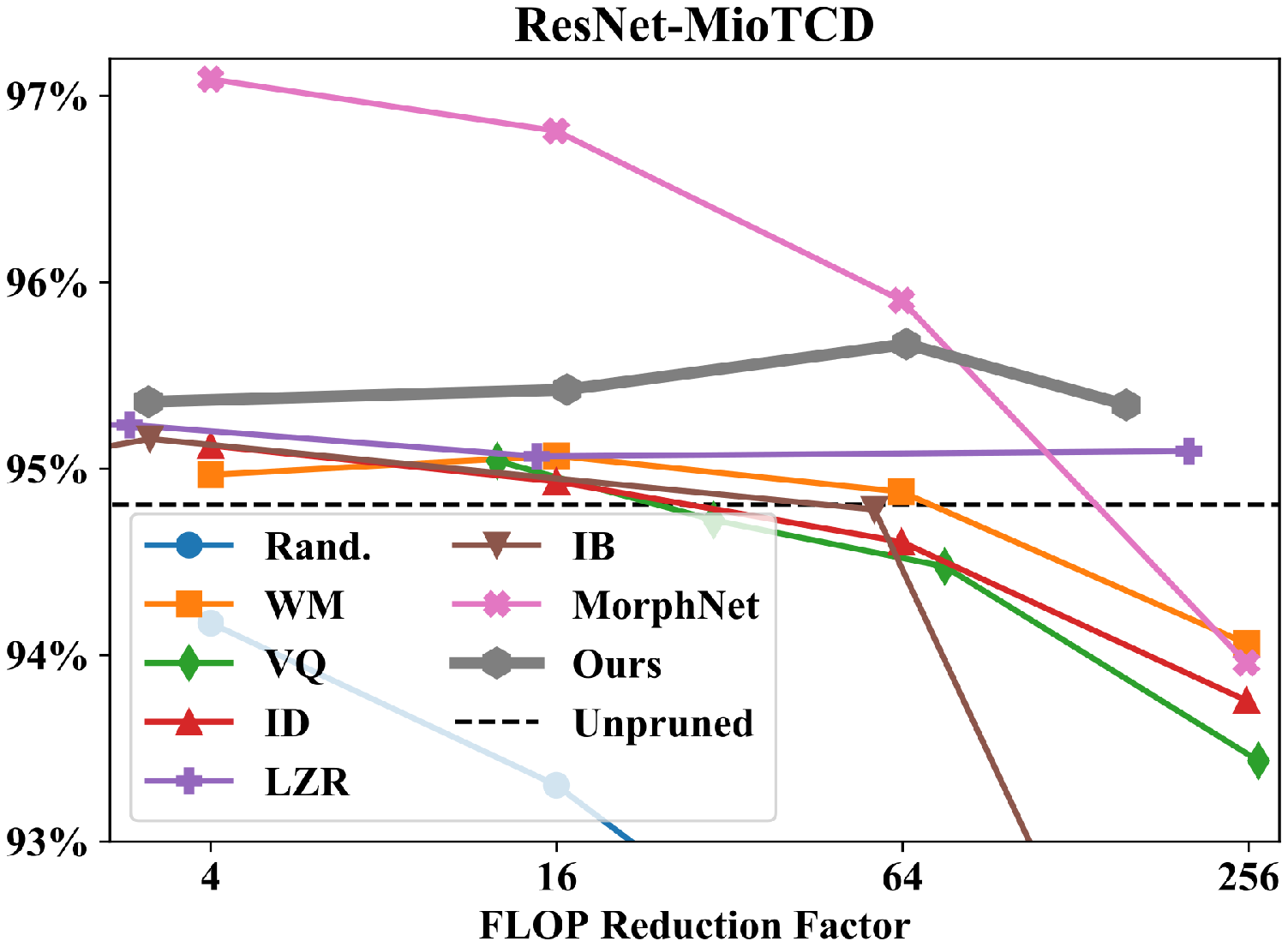}\vspace{-0.5cm}
\end{center}
   \caption{\textbf{Pruning results}. Plots showing test accuracy w.r.t. volume and FLOP reduction factor 
   (best viewed in color).}%\vspace{-0.2cm}
   \label{fig:results_plot}
\end{figure*}

Results for every method executed on all four datasets are shown in \figref{fig:results_plot}.  The first row shows test accuracies w.r.t. the network volume reduction factor for CIFAR-10, CIFAR-100, TinyImageNet and Mio-TCD.  As one can see, our method is above the others (or competitive) for CIFAR-10 and CIFAR-100.  It is also above every other method on TinyImageNet and Mio-TCD except for MorphNet which is better for pruning factors of 2 and 4.  However, MorphNet gets a severe drop of accuracy at 16x, a phenomena we observed as well on CIFAR-10 and CIFAR-100.  Our method is also always better than IB and LZR, the other two DSL methods.  Overall, our method is resilient to severe (16x) pruning ratios.

Furthermore, for every dataset, networks pruned with our method (as well as some others) get better results than the initial unpruned network.  This illustrates the fact that Wide-ResNet and ResNet-50 are overparameterized for certain tasks and that decreasing their number of feature maps reduces overfitting and thus improves test accuracy.

We then took every pruned network and computed their FLOP reduction factor (we considered operations from convolutions only).  This is illustrated in the second row of \figref{fig:results_plot}.  There again, our method outperforms (or is competitive with) the others for CIFAR-10 and CIFAR-100.  Our method reduces FLOPs by up to a factor of $\sim\!64$x on CIFAR-10, $\sim\!60$x on CIFAR-100 and $\sim\!200$x on Mio-TCD without decreasing test accuracy.  We get similar results as LZR for pruning ratios around $60$x on CIFAR-10 and CIFAR-100 and $200$x on Mio-TCD.  MorphNet gets better accuracy for pruning ratios of $4$x and $16$x on Mio-TCD, but then drops significantly around $256$x. Results are similar for TinyImageNet.

%\subsection{Ablation study}

%We also compare the performance of our method when removing some of its components. 
In \tabref{tab:ablation}, we report results of an ablation study on WideResNet-CIFAR-10 with two pruning factors. We replaced the Knowledge Distillation data loss (c.f. \secref{kds}) by a cross-entropy loss, and changed the Sigmoid pruning schedule (c.f. \secref{pruningtargettrans}) by a linear one. As can be seen, removing either of those reduces accuracy, thus showing their efficiency.  We also studied the impact of not using the mixed-connectivity block introduced in \secref{atypical}.  As shown in \tabref{tab:mixtConnBl}, when replacing our mixed-connectivity blocks by regular ResBlocks, we get a drop of the effective pruned volume of more than 50\% for 16x (even up to 58\% for CIFAR-10).

\begin{table}[tp]
\centering
\caption{\textbf{Test Accuracy for different configurations of our method (using WideResNet-CIFAR-10)}. The test accuracy of the unpruned network is $90.90\%$.}
\hspace{6em}
\label{tab:ablation}
\begin{tabular}{lcc}
\hline
\multirow{2}{*}{\textbf{Configuration}} & \multicolumn{2}{c}{\textbf{Pruning factor}} \\

                                & 2x                    & 16x                   \\ \hline
Our method                              & 92.70\%              & 91.62\%              \\
w/o Knowledge Distillation              & -1.37\%              & -0.40\%              \\
w/o Sigmoid pruning schedule            & -0.87\%              & -0.92\%              \\
\hline
\end{tabular}
\end{table}

\begin{table}[tp]
\centering
\caption{\textbf{Reduction of the effective pruned volume when removing the mixed-connectivity block.}}
\hspace{6em}
\label{tab:mixtConnBl}
\begin{tabular}{lcccc}
\hline
\textbf{Dataset}  &  2x   & 4x     & 8x    & 16x \\ \hline
CIFAR-10  & 12\% & 43\% & 53\% & 58\% \\
CIFAR-100 & 14\% & 49\% & 55\% & 57\% \\
MIO-TCD   & 32\% & 37\% & 40\% & 52\% \\
\hline
\end{tabular}\vspace{-0.0cm}
\end{table}

We illustrate in \figref{fig:layerpruning} results of our pruning method for CIFAR-10 (for the other datasets, see supplementary materials).  The figure shows the number of neurons per residual block for the full network, and for the networks pruned with varying pruning factors.  These plots show that our method has the capability of eliminating entire residual blocks (especially around 1.3 and 1.4).  Also, the pruning configurations follow no obvious trend thus showing the inherent plasticity of a DSL method such as ours.

% \begin{figure*}[tp]
%   \centering
%   \subfloat[WideResNet-CIFAR-10]{\includegraphics[width=0.335\linewidth]{bars_c10.svg.eps}}
%   \hspace{1pt}
%   \subfloat[WideResNet-CIFAR-100]{\includegraphics[width=0.325\linewidth]{bars_c100.svg.eps}}
%   \hspace{1pt}
%   \subfloat[ResNet-Mio-TCD]{\includegraphics[width=0.325\linewidth]{bars_tcd.svg.eps}}
%   \caption{\textbf{Result of pruning with our method}. Total number of active neurons in the full networks and with four different compression rates.  Sections without an orange (8x) or red (16x) bar are those for which a res-Block has been eliminated.}
%   \label{fig:layerpruning}
% \end{figure*}

\begin{figure}
  \centering
  \includegraphics[width=0.99\linewidth]{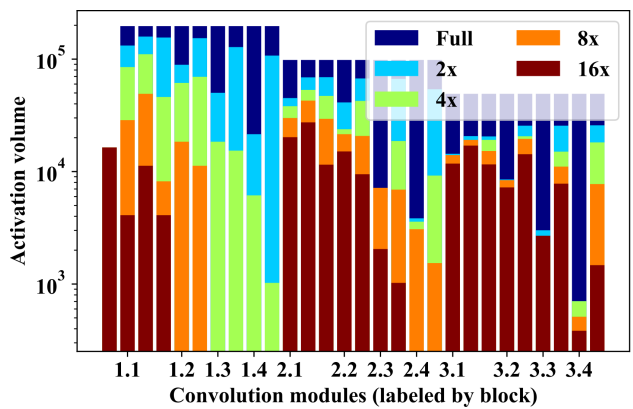}
  \caption{\textbf{Result of pruning with our method} on WideResNet-CIFAR-10. Total number of active neurons in the full networks and with four different pruning rates.  Sections without an orange (8x) or red (16x) bar are those for which a res-Block has been eliminated.}
  \label{fig:layerpruning}
\end{figure}

\begin{figure}[tp]
\vspace{-1em}
  \centering
  {\includegraphics[width=0.48\linewidth]{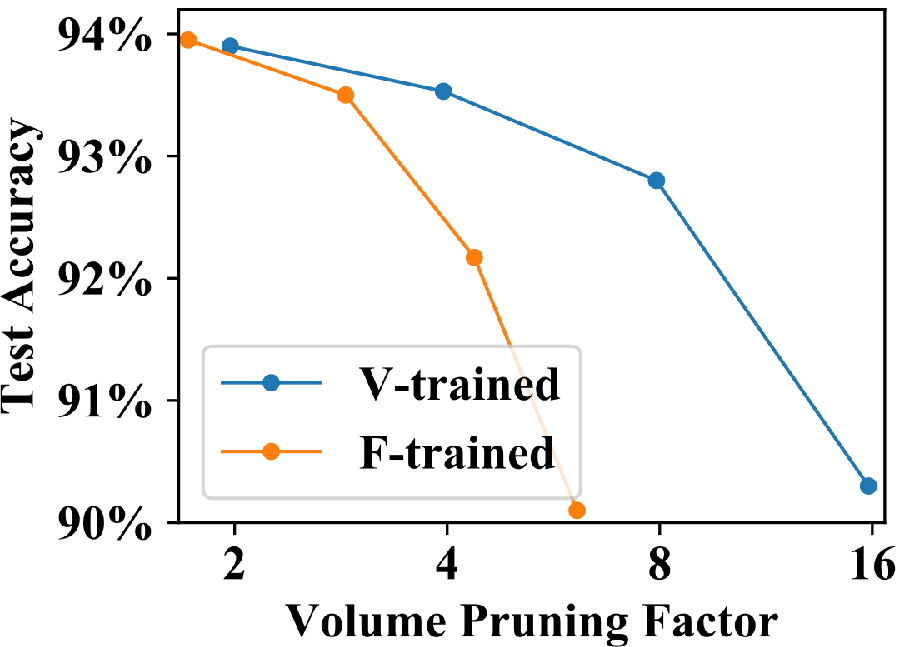}}
  %\hspace{2pt}
  {\includegraphics[width=0.48\linewidth]{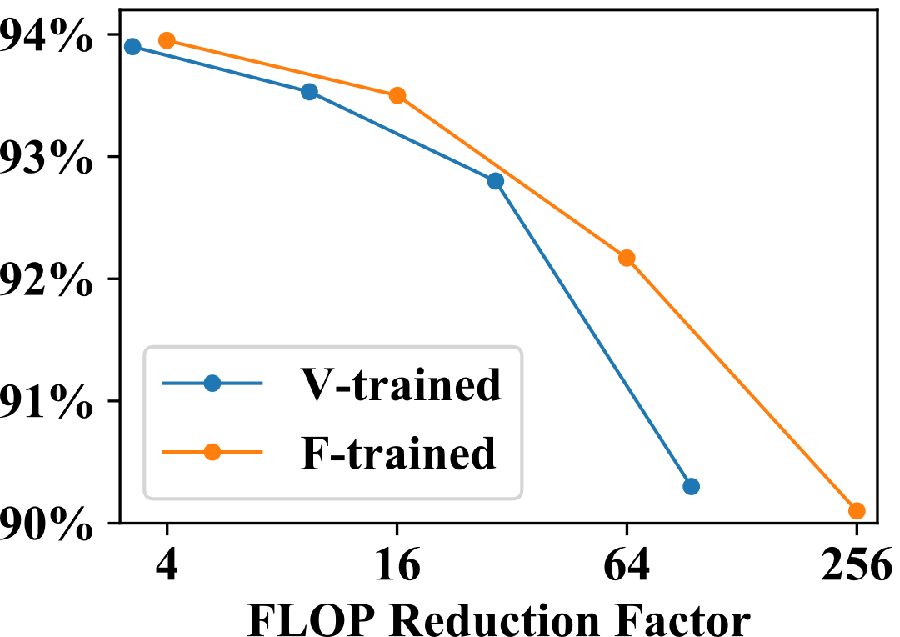}}\vspace{-0.2cm}
   \caption{\textbf{Comparison of objective metrics}. Test accuracy versus the volume pruning factor and the FLOP reduction factor for our method with a Volume metric (V-trained) and a FLOP metric (F-trained).\vspace{-0.5cm}}
   \label{fig:vftrained}
\end{figure}

As mentioned in \secref{bar}, instead of the volume metric  (\eqref{eq:omega}) the budget could be set w.r.t a FLOP metric by accounting for the expectation of the number of feature maps in the preceding layer. We compare in \figref{fig:vftrained} the results given by these two budget metrics for WideResnet-CIFAR-10.  As one might expect, pruning a network with a volume metric ({\em V-Trained}) yields significantly better performances w.r.t. the volume pruning factor whereas pruning a network with a FLOP metric ({\em F-Trained}) yields better performances w.r.t. to the FLOP reduction factor, although by a slight margin.  In light of these results, we conclude that the volume metric (\eqref{eq:omega}) is overall a better choice.

\vspace{-0.1cm}
\section{Conclusion}
\vspace{-0.1cm}
We presented a structured budgeted pruning method based on a dropout sparsity learning framework. We proposed a knowledge distillation loss function combined with a budget-constrained sparsity loss whose formulation is that of a barrier function.  Since the log-barrier solution is ill-suited for pruning a CNN, we proposed a novel barrier function as well as a novel optimization schedule.  We provided concrete insights on how to prune residual networks and used a novel mixed-connectivity block. Results obtained on two ResNets architecture and four datasets reveal that our method outperforms (or is competitive to) 7 other pruning methods.

\vspace{-0.2cm}
\section*{Acknowledgements}
\vspace{-0.2cm}

{\footnotesize We thank Christian Desrosiers for his insights. This work was supported by FRQ-NT scholarship \#257800 and Mitacs grant IT08995. Supercomputers from Compute Canada and Calcul Quebec were used.}

{\footnotesize
\bibliographystyle{plain}
\bibliography{biblio}
}

\end{document}

%%%%%%%%% TITLE
\setlength{\droptitle}{-1cm}

\title{Supplementary Materials for \\ ``Structured Pruning of Neural Networks \\ with Budget-Aware Regularization"}
\date{}

\maketitle
\vspace{-1.5cm}

%\tableofcontents
%\vspace{1cm}

\section{Sigmoidal Transition Function} \label{app:sigmoid}

The sigmoidal transition function of Section 3.4 is defined as follows:

\begin{equation} \label{eq:sigmoidal}
\begin{split}
    T(t, d) &= \frac{\mathrm{Sigmoid}(d(t-0.5))-\delta}{1-2\delta} \\
    \delta &= \mathrm{Sigmoid}(-0.5d)
\end{split}
\end{equation}

where $d$ controls the ``hardness" of the sigmoid curve. $d \rightarrow 0$ gives a linear transition, and $d \rightarrow \infty$ gives a hard sigmoid. $d=10$ was found empirically to be effective. The sigmoid must be shifted by $\delta$ and scaled by $(1-2\delta)$ so that $T(0,d)=0$ and $T(1,d)=1$.% We set $d=10$ for all our experiments.

\section{Mixed Block Implementation} \label{appen:mixedblock}

In Algorithm \ref{algo:mixedblock} below, we present an implementation for the atypical connectivity presented in Figure 4 of the paper. While the input and delta tensors have the same number of prunable features, they can have a different pruning mask. We use indexing operations to exclude pruned features from the computations. 
%We feed the alive features of the residual signal into the sequence of convolutions $f_\Delta(\cdot)$ of the block. The resulting features $\Delta$ are mapped to their corresponding features in the residual signal using the indices $i_\Delta$ to allow summing.

Note that, in the case where the block is responsible of performing a downsampling of the input tensor (or augmenting the number of features, or both), the residual connection is not an identity function; rather, it is a convolution $f_\mathrm{res}(\cdot)$ that can have pruned features.

\begin{algorithm}[H]
    \caption{Forward Pass for the Mixed Block}\label{algo:mixedblock}
    \DontPrintSemicolon
    \KwData{\\
    $x_\mathrm{in}$ : Input tensor of size $(\mathrm{batch\_size},\mathrm{n\_features},\mathrm{height},\mathrm{width})$ \\
    $f_\Delta(\cdot)$ : Delta branch function \\
    $i_\mathrm{in}$ : Indices of alive features in input tensor \\
    $i_\Delta$ : Indices of alive features in delta tensor \\
    $f_\mathrm{res}(\cdot)$ : Residual branch function (can be $\varnothing$ if identity connection) \\
    $i_\mathrm{res}$ : Indices of alive features in output of residual function (if $f_\mathrm{res}(\cdot) \neq \varnothing$) \\
    $N_\mathrm{res}$ : Number of output features of unpruned residual function (if $f_\mathrm{res}(\cdot) \neq \varnothing$) \\
    }
    \KwResult{$x_\mathrm{out}$ : Output tensor}
    \BlankLine
    
    $x_\mathrm{alive} \Leftarrow x_\mathrm{in}[:,i_\mathrm{in},:,:]$ \;
    %$x_\Delta \Leftarrow f_\Delta(x_\mathrm{alive})$ \;
    
    % \mathrm{\ is\ set}$
    \If{$f_\mathrm{res}(\cdot) = \varnothing$}{
        $x_\mathrm{out} \Leftarrow i_\mathrm{in}$ \;
    }
    \Else{
        $x_\mathrm{res} \Leftarrow f_\mathrm{res}(x_\mathrm{alive})$ \;
        $\mathrm{height},\mathrm{width} \Leftarrow \mathrm{get\_height}(x_\mathrm{res}),\mathrm{get\_width}(x_\mathrm{res})$ \;
        $x_\mathrm{out} \Leftarrow \mathrm{zeros}(\mathrm{batch\_size},N_\mathrm{res},\mathrm{height},\mathrm{width})$ \;
        $x_\mathrm{out}[:,i_\mathrm{res},:,:] \Leftarrow x_\mathrm{res}$ \;
    }
    
    $\Delta \Leftarrow f_\Delta(x_\mathrm{alive})$ \;
    $x_\mathrm{out}[:,i_\Delta,:,:] \Leftarrow x_\mathrm{out}[:,i_\Delta,:,:] + \Delta$ \;
\end{algorithm}

% \begin{equation*}
% \begin{split}
%     \Delta &= f(h_{l-1}[I_\mathrm{in}]) \\
%     h_l &= \texttt{scatterAdd}(h_{l-1}, \Delta, I_\mathrm{out}).
% \end{split}
% \end{equation*}
% The output of the block is $h_l$, which can be viewed as the ``refined residual signal". Here, \texttt{scatterAdd} creates a tensor $h_l$ where $h_l[I_\mathrm{out}] = h_{l-1}[I_\mathrm{out}] + \Delta$ and $h_l[\neg I_\mathrm{out}] = h_{l-1}[\neg I_\mathrm{out}]$. Scattering operations are available in deep learning libraries such as \textit{PyTorch} or \textit{TensorFlow}. A visual explanation is available in the 
% \href{https://www.tensorflow.org/api_docs/python/tf/scatter_add}{documentation of TensorFlow's \texttt{scatter\_add}}.

\section{Training Schedules}

Here, we give details about the number of epochs used for each training phase and for each method.

\subsection{Training Schedule for LZR, IB, MorphNet and Our Method}

\noindent
\textbf{LZR[22], IB[1] and Ours} \indent For CIFAR-10 and CIFAR-100, we first train for 80 epochs. Then, the network is ``hard-pruned" using the algorithm associated to the pruning method. Next, we freeze $\Phi$, and we train for 40 additional epochs. Finally, we reduce the learning rate from $10^{-3}$ to $10^{-4}$ and train for 10 more epochs.  For Mio-TCD, we shorten the three stages to $(40, 20, 10)$ epochs (this dataset is $\sim 10 \times$ larger). For Mio-TCD only, we initialize the weights to those of the full (unpruned) network (c.f. \secref{unpruned}).

\vspace{1em} \noindent
\textbf{MorphNet[9]} \indent For CIFAR-10 and CIFAR-100, we first train for 60 epochs. Then, the network is ``hard-pruned" using the algorithm proposed by the original paper. Next, we train for 50 additional epochs. Finally, we reduce the learning rate from $10^{-3}$ to $10^{-4}$ and train for 20 more epochs. For Mio-TCD, we shorten the three stages to $(40, 20, 10)$ epochs.

%\begin{samepage}
\subsection{Pruning Scheme for Random, VM, VQ and ID}
For {\em Random}, {\em WM}[10], {\em VQ}[8], and {\em ID}[6,16], we implemented the following pruning scheme which revealed to be effective and efficient:

\begin{enumerate}
    \item Perform initial training of the full network for 40 epochs; 
    \item Reduce learning rate and continue training for 10 epochs; 
    \item \label{enum:remove} Prune 50\% of the network's volume $V$;
    \item Train the network again with high learning rate, for 40 epochs;
    \item Reduce learning rate and continue training for 10 epochs; 
    \item Record the network performance for the current pruning factor;
    \item Return to \ref*{enum:remove} for the remaining pruning steps.
\end{enumerate}

This scheme leads to four pruning factors: 2, 4, 8, 16. %The results of our experiments on CIFAR-10, CIFAR-100 and MIO-TCD are plotted in Figs. \ref{fig:results_CIFAR-10}, \ref{fig:results_CIFAR-100} and \ref{fig:results_tcd}, respectively.
%\end{samepage}

\subsection{Training Schedule for the Full (Unpruned) Networks} \label{unpruned}

For CIFAR-10 and CIFAR-100, we trained for 80 epochs at learning rate $10^{-3}$, and 10 more epochs at learning rate $10^{-4}$. For Mio-TCD, a dataset $\sim10 \times$ larger, we trained for 40 epochs at high learning rate and 10 epochs at low learning rate.

\section{Initialization of Dropout Sparsity Parameters $\Phi$}

For LZR, where $\Phi := \{\alpha_l\}$, we initialized all $\alpha$ from a uniform distribution $\mathcal{U}(0, 0.01)$. We did not observe a significant change between using $\mathcal{U}(0, 0.01)$ or $\mathcal{N}(0, 0.01^2)$ (the initialization distribution suggested by [22]). Our method has the same $\Phi$ and initialization scheme than LZR[22]. For IB, where $\Phi := \{\mu_l,\sigma_l\}$, we initialized the parameters from Gaussian distributions: $\mu \sim \mathcal{N}(1, 0.01^2), \log \sigma \sim \mathcal{N}(-9, 0.01^2)$ (values obtained by personal communication with the authors~[1]).

\section{Impact of Mixed-Connectivity Block on Metrics}

Here we compare the objective metrics (i.e. Activation Volume $V$ and FLOP) when the regular Resblock is used and when our Mixed-Connectivity block is used. These are the results obtained from our method (BAR). $\Delta$ is the relative difference to the results for the Regular block.  The following three tables provide further details to Table 2 in the paper.

\subsection{CIFAR-10}
\begin{table}[H]
\begin{tabular}{c|rrr|rrr}
\hline
               & \multicolumn{3}{c|}{Activation Volume ($V$)} & \multicolumn{3}{c}{FLOP}       \\
Pruning Factor & Regular       & Mixed         & $\Delta$     & Regular  & Mixed    & $\Delta$ \\ \hline
2              & 1.79E+06      & 1.58E+06      & -12\%        & 2.71E+09 & 2.35E+09 & -13\%    \\
4              & 1.37E+06      & 7.88E+05      & -43\%        & 1.51E+09 & 9.49E+08 & -37\%    \\
8              & 8.43E+05      & 3.93E+05      & -53\%        & 8.06E+08 & 4.46E+08 & -45\%    \\
16             & 4.67E+05      & 1.97E+05      & -58\%        & 4.20E+08 & 2.20E+08 & -48\%    \\ \hline
\end{tabular}
\end{table}

\subsection{CIFAR-100}
\begin{table}[H]
\begin{tabular}{c|rrr|rrr}
\hline
               & \multicolumn{3}{c|}{Activation Volume ($V$)} & \multicolumn{3}{c}{FLOP}       \\
Pruning Factor & Regular       & Mixed         & $\Delta$     & Regular  & Mixed    & $\Delta$ \\ \hline
2              & 1.84E+06      & 1.58E+06      & -14\%        & 2.68E+09 & 2.29E+09 & -15\%    \\
4              & 1.53E+06      & 7.88E+05      & -49\%        & 1.85E+09 & 1.10E+09 & -40\%    \\
8              & 8.76E+05      & 3.93E+05      & -55\%        & 1.08E+09 & 6.03E+08 & -44\%    \\
16             & 4.59E+05      & 1.97E+05      & -57\%        & 6.21E+08 & 3.27E+08 & -47\%    \\ \hline
\end{tabular}
\end{table}

\subsection{Mio-TCD}
\begin{table}[H]
\begin{tabular}{c|rrr|rrr}
\hline
               & \multicolumn{3}{c|}{Activation Volume ($V$)} & \multicolumn{3}{c}{FLOP}       \\
Pruning Factor & Regular       & Mixed         & $\Delta$     & Regular  & Mixed    & $\Delta$ \\ \hline
2              & 2.66E+06      & 1.81E+06      & -32\%        & 1.15E+09 & 8.33E+08 & -28\%    \\
4              & 1.45E+06      & 9.07E+05      & -37\%        & 2.50E+08 & 1.56E+08 & -38\%    \\
8              & 7.51E+05      & 4.54E+05      & -40\%        & 6.87E+07 & 3.98E+07 & -42\%    \\
16             & 4.70E+05      & 2.26E+05      & -52\%        & 3.31E+07 & 1.66E+07 & -50\%    \\ \hline
\end{tabular}
\end{table}

\section{Pruning Results}

Here we give the numbers that we used to plot the curves of Figures 6 and 8. All methods prune according to the Activation Volume $V$, except for ``Ours (F-trained)", which prunes with a FLOP reduction objective. Please note that results with a volume factor greater than 16 have not been included.

\subsection{CIFAR-10}

\begin{table}[H]
\begin{tabular}{lrrr|lrrr}
\hline
Method & V factor & F factor & Test Accu. & Method      & V factor & F factor & Test Accu. \\ \hline
Random & 2.0      & 4.0      & 0.8963     & LZR         & 1.1      & 1.2      & 0.9182     \\
       & 4.0      & 16.0     & 0.8802     &             & 2.1      & 5.3      & 0.9204     \\
       & 8.0      & 64.0     & 0.8462     &             & 4.3      & 38.1     & 0.9210     \\
       & 16.0     & 255.8    & 0.8136     &             & 7.3      & 62.0     & 0.9171     \\ \cline{1-4}
VQ     & 3.1      & 9.4      & 0.9124     &             & 15.8     & 167.4    & 0.8970     \\ \cline{5-8} 
       & 4.6      & 21.0     & 0.9089     & IB          & 1.3      & 1.8      & 0.9085     \\
       & 8.1      & 64.9     & 0.8862     &             & 1.3      & 1.8      & 0.9128     \\
       & 16.0     & 256.7    & 0.8459     &             & 2.3      & 5.2      & 0.9097     \\ \cline{1-4}
WM     & 2.0      & 4.0      & 0.9111     &             & 5.8      & 38.6     & 0.9048     \\
       & 4.0      & 16.0     & 0.9120     &             & 7.4      & 79.9     & 0.9014     \\
       & 8.0      & 64.0     & 0.8968     &             & 9.5      & 113.7    & 0.8965     \\
       & 16.0     & 255.8    & 0.8633     &             & 15.2     & 196.4    & 0.8795     \\ \hline
ID     & 2.0      & 4.0      & 0.9109     & MorphNet    & 2.0      & 4.0      & 0.9325     \\
       & 4.0      & 16.0     & 0.9144     &             & 4.0      & 16.0     & 0.9289     \\
       & 8.0      & 64.0     & 0.9037     &             & \dag~7.9      & 64.0     & 0.9066     \\
       & 16.0     & 255.8    & 0.8692     &             & \dag~15.8     & 255.8    & 0.8643     \\ \hline
Ours   & 2.0      & 6.4      & 0.9270     & Ours        & 1.7      & 4.0      & 0.9395     \\
       & 4.0      & 15.9     & 0.9278     & (F-trained) & 2.9      & 16.0     & 0.9350     \\
       & \dag~7.9      & 33.8     & 0.9280     &             & 4.4      & \dag~63.9     & 0.9217     \\
       & \dag~15.8     & 68.5     & 0.9162     &             & 6.1      & 256.0    & 0.9010     \\ \hline
\end{tabular}
\vspace{1em} \\
\footnotesize{The results marked with a dagger (\dag) are slightly below target because of a slight miscalculation of the total volume and of the budget; this has since been fixed in our implementation.}
\end{table}

\subsection{CIFAR-100}

\begin{table}[H]
\begin{tabular}{lrrr|lrrr}
\hline
Method & V factor & F factor & Test Accu. & Method   & V factor & F factor & Test Accu. \\ \hline
Random & 2.0      & 4.0      & 0.6751     & LZR      & 1.0      & 1.1      & 0.7201     \\
       & 4.0      & 16.0     & 0.6364     &          & 2.0      & 6.0      & 0.7039     \\
       & 8.0      & 64.0     & 0.5922     &          & 3.0      & 31.9     & 0.7152     \\
       & 16.0     & 255.8    & 0.4888     &          & 6.6      & 57.5     & 0.7075     \\ \cline{1-4}
VQ     & 2.5      & 9.4      & 0.6963     &          & 13.0     & 110.6    & 0.6728     \\ \cline{5-8} 
       & 4.3      & 21.0     & 0.6927     & IB       & 1.2      & 1.5      & 0.7127     \\
       & 8.0      & 64.9     & 0.6703     &          & 1.3      & 1.8      & 0.7093     \\
       & 16.0     & 256.7    & 0.5899     &          & 2.0      & 3.1      & 0.7079     \\ \cline{1-4}
WM     & 2.0      & 4.0      & 0.6898     &          & 4.8      & 24.9     & 0.6508     \\
       & 4.0      & 16.0     & 0.6910     &          & 6.3      & 58.5     & 0.6905     \\
       & 8.0      & 64.0     & 0.6542     &          & 8.8      & 94.3     & 0.5709     \\
       & 16.0     & 255.8    & 0.5899     &          & 12.7     & 148.5    & 0.5671     \\ \hline
ID     & 2.0      & 4.0      & 0.6929     & MorphNet & 2.0      & 4.0      & 0.7359     \\
       & 4.0      & 16.0     & 0.6975     &          & 4.0      & 16.0     & 0.7042     \\
       & 8.0      & 64.0     & 0.6603     &          & \dag~7.9      & 64.0     & 0.6494     \\
       & 16.0     & 255.8    & 0.5913     &          & \dag~15.8     & 255.8    & 0.5549     \\ \cline{1-4}
Ours   & 2.0      & 6.6      & 0.7408     &          &          &          &            \\
       & 4.0      & 13.7     & 0.7359     &          &          &          &            \\
       & \dag~7.9      & 25.0     & 0.7259     &          &          &          &            \\
       & \dag~15.8     & 46.2     & 0.7053     &          &          &          &            \\ \hline
\end{tabular}
\vspace{1em} \\
\footnotesize{The dagger (\dag) has the same meaning as in the previous table.}
\end{table}

\subsection{Mio-TCD}

\begin{table}[H]
\begin{tabular}{lrrr|lrrr}
\hline
Method & V factor & F factor & Test Accu. & Method   & V factor & F factor & Test Accu. \\ \hline
Random & 2.0      & 4.0      & 0.9417     & LZR      & 1.0      & 1.0      & 0.9521     \\
       & 4.0      & 16.0     & 0.9330     &          & 1.0      & 1.0      & 0.9509     \\
       & 8.0      & 63.7     & 0.9191     &          & 1.0      & 1.0      & 0.9524     \\
       & 16.0     & 253.7    & 0.9119     &          & 1.6      & 2.9      & 0.9524     \\ \cline{1-4}
VQ     & 3.6      & 12.6     & 0.9504     &          & 3.8      & 14.8     & 0.9507     \\
       & 5.5      & 30.0     & 0.9472     &          & 11.7     & 201.6    & 0.9510     \\ \cline{5-8} 
       & 8.7      & 75.8     & 0.9448     & IB       & 1.4      & 2.3      & 0.9509     \\
       & 16.3     & 266.3    & 0.9343     &          & 1.9      & 3.1      & 0.9516     \\ \cline{1-4}
WM     & 2.0      & 4.0      & 0.9497     &          & 7.5      & 57.3     & 0.9478     \\
       & 4.0      & 16.0     & 0.9507     &          & 11.5     & 145.0    & 0.9211     \\
       & 8.0      & 63.7     & 0.9488     &          & 14.2     & 192.6    & 0.9127     \\
       & 16.0     & 253.7    & 0.9406     &          & 15.2     & 232.6    & 0.9070     \\ \hline
ID     & 2.0      & 4.0      & 0.9512     & MorphNet & 2.0      & 4.0      & 0.9709     \\
       & 4.0      & 16.0     & 0.9493     &          & 4.0      & 16.0     & 0.9681     \\
       & 8.0      & 63.7     & 0.9461     &          & 8.0      & 63.7     & 0.9590     \\
       & 16.0     & 253.7    & 0.9376     &          & 16.0     & 253.7    & 0.9396     \\ \cline{1-4}
Ours   & 2.0      & 3.1      & 0.9536     &          &          &          &            \\
       & 4.0      & 16.7     & 0.9543     &          &          &          &            \\
       & 8.0      & 65.1     & 0.9567     &          &          &          &            \\
       & 16.1     & 156.7    & 0.9534     &          &          &          &            \\ \hline
\end{tabular}
\end{table}

\subsection{TinyImageNet}

\begin{table}[H]
\begin{tabular}{lrrr|lrrr}
\hline
Method & V factor & F factor & Test Accu. & Method   & V factor & F factor & Test Accu. \\ \hline
Random & 2.0      & 4.0      & 0.4825     & Ours     & 2.0      & 4.0      & 0.5235     \\
       & 4.0      & 16.0     & 0.4608     &          & 4.0      & 16.0     & 0.5198     \\
       & 8.0      & 63.7     & 0.3941     &          & 8.0      & 63.7     & 0.5140     \\
       & 16.0     & 253.7    & 0.2953     &          & 16.0     & 253.7    & 0.5196     \\ \hline
VQ     & 2.5      & 6.2      & 0.4896     & LZR      & 3.4      & 38.8     & 0.5034     \\
       & 4.2      & 18.0     & 0.4993     &          & 4.0      & 55.3     & 0.5001     \\
       & 8.0      & 64.3     & 0.4842     &          & 5.5      & 73.8     & 0.4971     \\
       & 16.0     & 255.9    & 0.3926     &          & 9.6      & 108.3    & 0.4903     \\ \hline
WM     & 2.0      & 4.0      & 0.4901     & IB       & 2.0      & 3.9      & 0.4445     \\
       & 4.0      & 16.0     & 0.4967     &          & 4.2      & 18.0     & 0.4282     \\
       & 8.0      & 63.7     & 0.4772     &          & 6.4      & 41.3     & 0.3757     \\
       & 16.0     & 253.7    & 0.4019     &          & 12.7     & 161.7    & 0.3513     \\ \hline
ID     & 2.0      & 4.0      & 0.4996     & MorphNet & 2.0      & 4.0      & 0.5815     \\
       & 4.0      & 16.0     & 0.4955     &          & 4.0      & 16.0     & 0.5577     \\
       & 8.0      & 63.7     & 0.4577     &          & 8.0      & 63.7     & 0.5169     \\
       & 16.0     & 253.7    & 0.3972     &          & 16.0     & 253.7    & 0.3919     \\ \hline
\end{tabular}
\end{table}

\section{Pruning Results}

\begin{figure}[H]
  \centering
  \subfloat[WideResNet-TinyImageNet]{\includegraphics[width=0.32\linewidth]{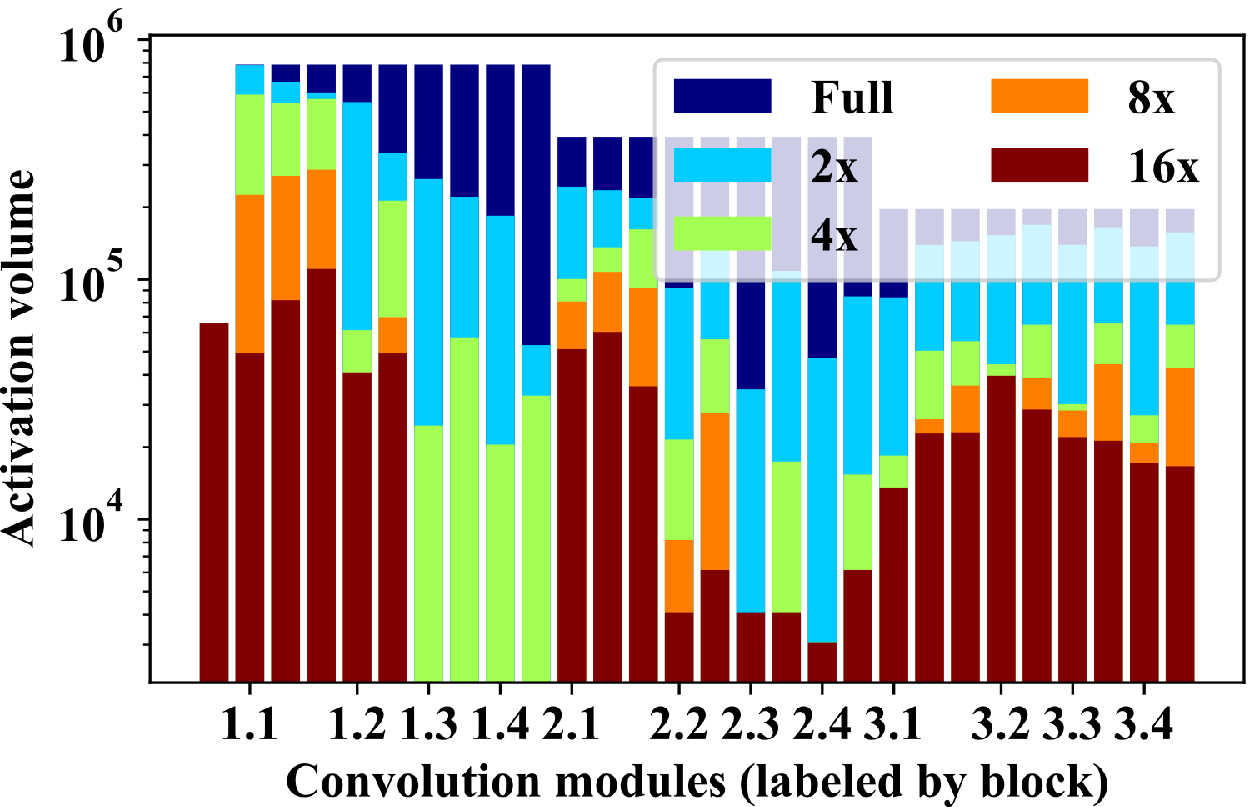}}
  \hspace{1pt}
  \subfloat[WideResNet-CIFAR-100]{\includegraphics[width=0.3\linewidth]{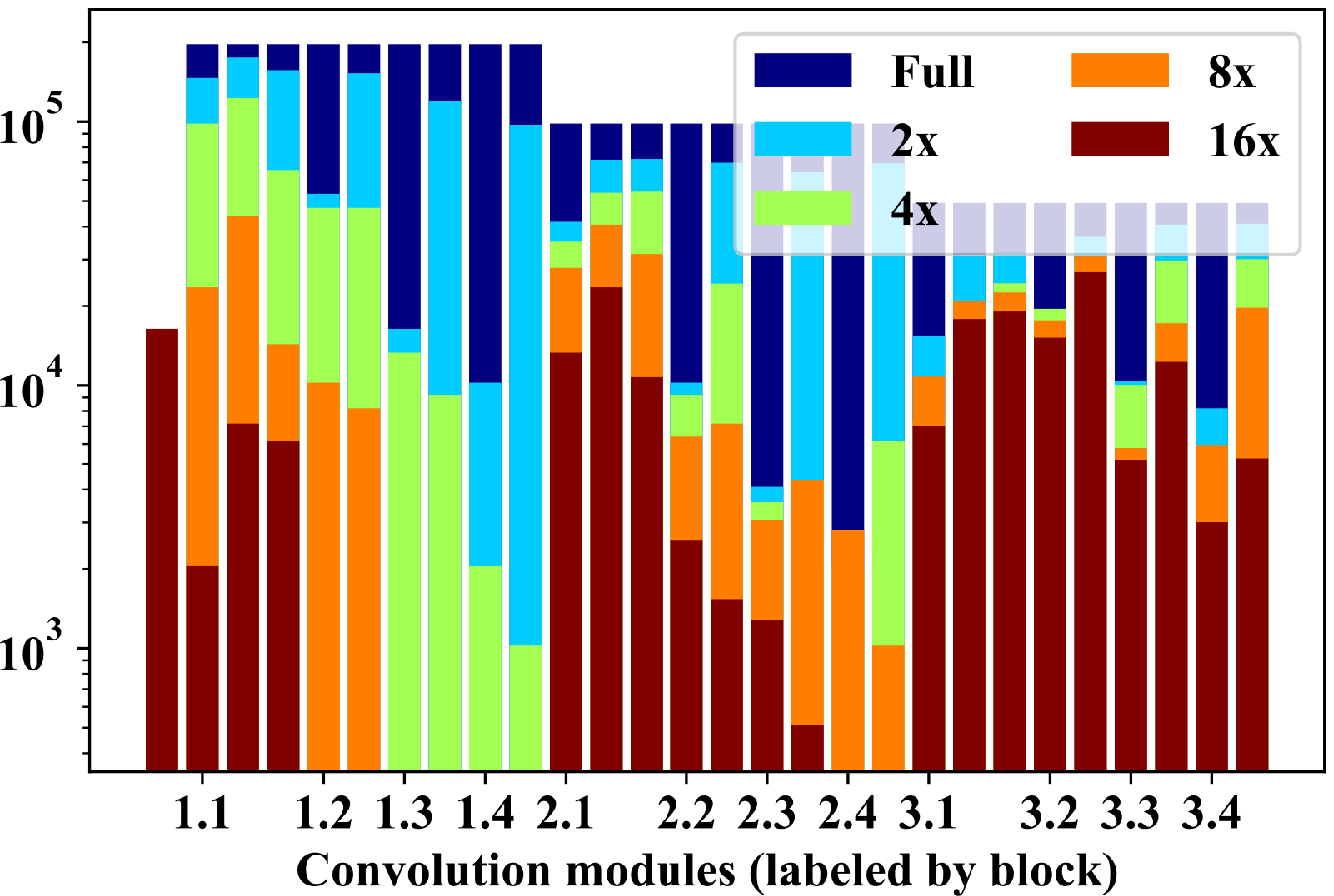}}
  \hspace{1pt}
  \subfloat[ResNet-Mio-TCD]{\includegraphics[width=0.3\linewidth]{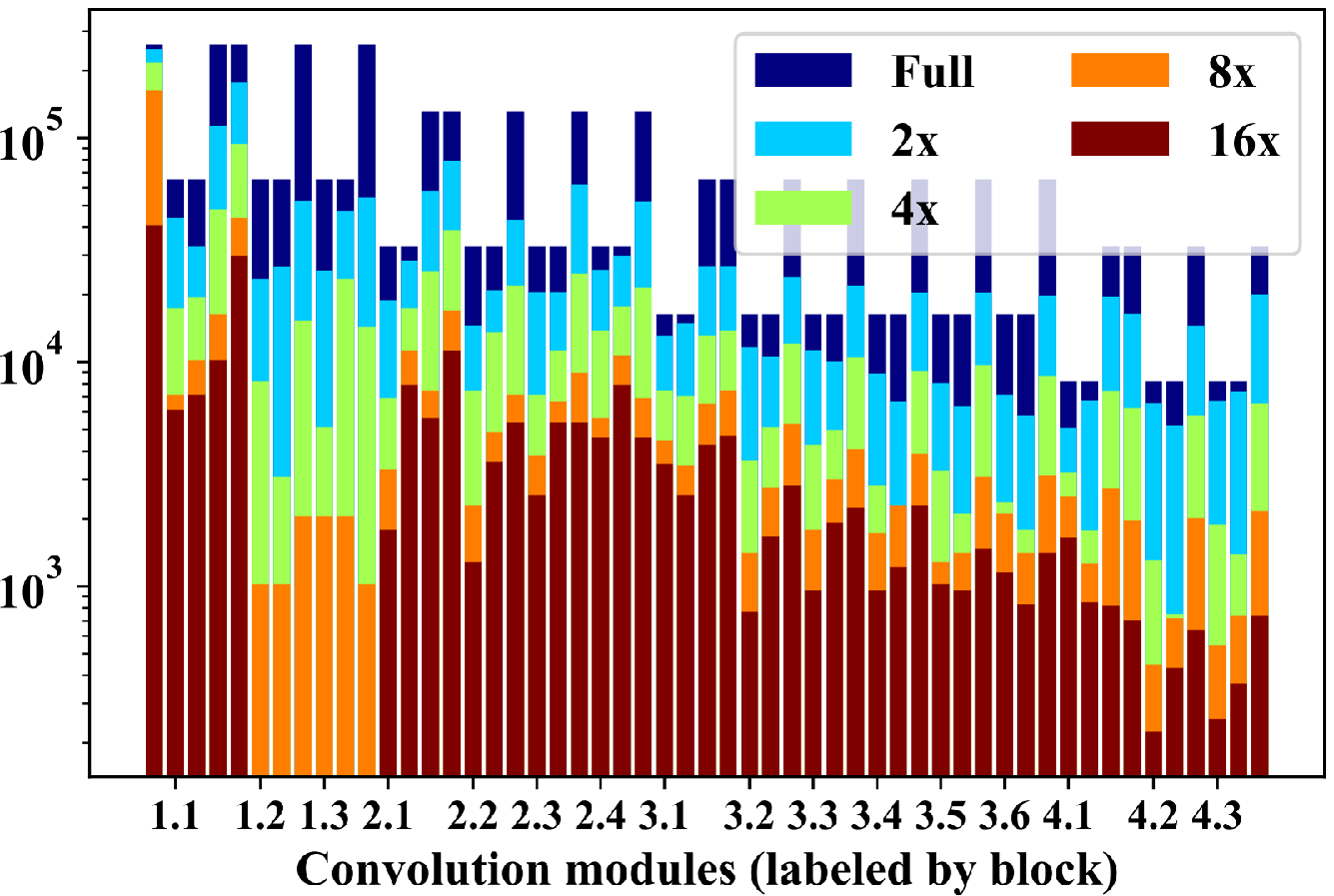}}
  \caption{\textbf{Result of pruning with our method}. Total number of active neurons in the full networks and with four different pruning rates.  Sections without an orange (8x) or red (16x) bar are those for which a res-Block has been eliminated.}
  \label{fig:layerpruning}
\end{figure}

In \figref{fig:layerpruning}, we show pruning results of our method on TinyImageNet, CIFAR-100, and Mio-TCD (result on CIFAR-10 is in the paper).

\section{Sensitivity Analysis}

\begin{figure}[H]
  \centering
  \includegraphics[width=\textwidth]{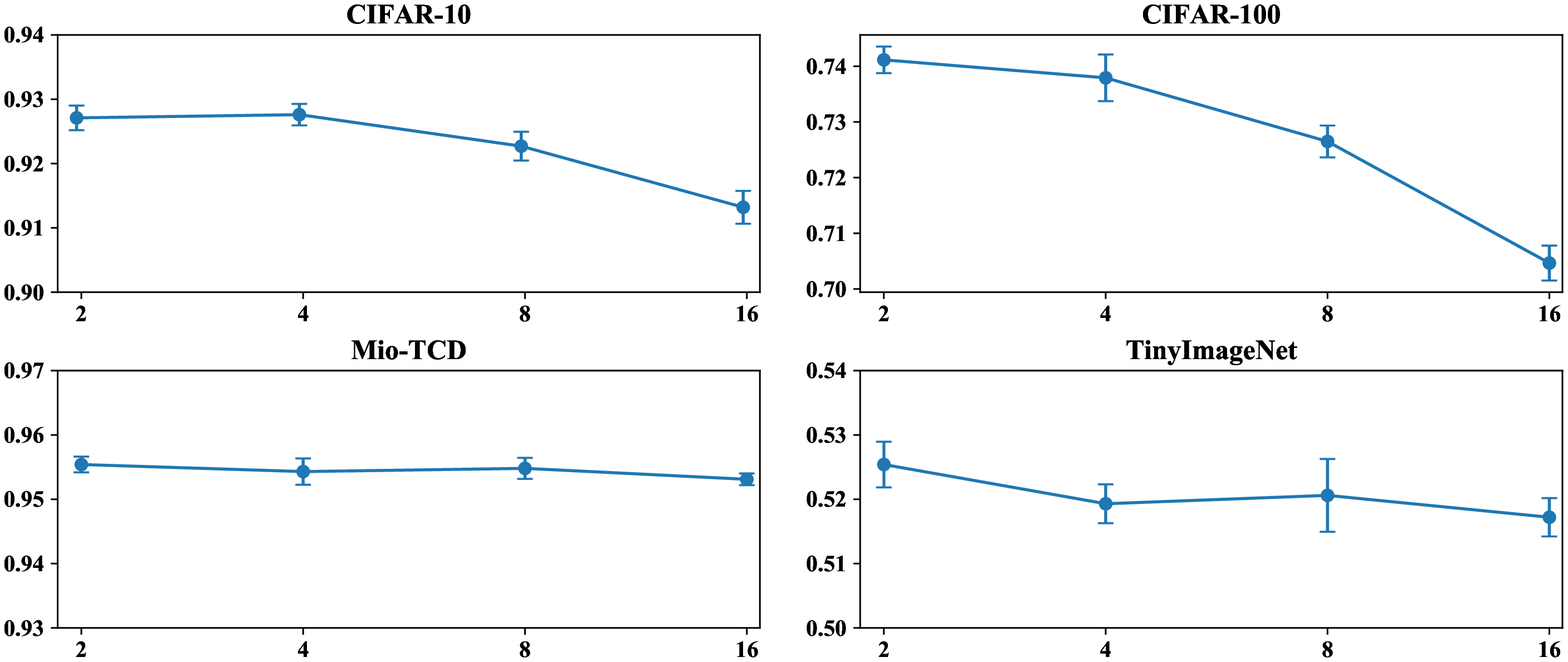}
  \caption{Sensitivity analysis with error bars}
  \label{sens-plot}
\end{figure}

Here we show sensitivity analysis results for our method on all 4 datasets. For each of the 4 pruning factors, we run 10 experiments, where we sample the value of 7 hyperparameters from a uniform distribution centered around their tuned value. Depending on the scale of the parameter, the width of the interval is either $10^{-5}$, $0.2$, or $1.0$. We have plotted the results of this analysis in \figref{sens-plot}, using the data shown in \tabref{sens-data}.  As one can see, our method is not over sensitive to changes of its hyperparameters.

\begin{table}[H]
    \centering
    \subfloat[][CIFAR-10]{
        \begin{tabular}{rrr}
        \hline
        V Factor & Test Accuracy & Std dev. \\ \hline
        2             & 0.9271        & 0.001915           \\
        4             & 0.9276        & 0.001674           \\
        8             & 0.9227        & 0.002244           \\
        16            & 0.9132        & 0.002552           \\ \hline
        \end{tabular}
    }
    ~ %add desired spacing between images, e. g. ~, \quad, \qquad, \hfill etc. 
    \subfloat[][CIFAR-100]{
        \begin{tabular}{rrr}
        \hline
        V Factor & Test Accuracy & Std dev. \\ \hline
        2             & 0.7412        & 0.002384           \\
        4             & 0.7379        & 0.004185           \\
        8             & 0.7265        & 0.002860           \\
        16            & 0.7047        & 0.003129           \\ \hline
        \end{tabular}
    }
    \\
    \subfloat[][Mio-TCD]{
        \begin{tabular}{rrr}
        \hline
        V Factor & Test Accuracy & Std dev. \\ \hline
        2             & 0.9554        & 0.001222           \\
        4             & 0.9543        & 0.002039           \\
        8             & 0.9548        & 0.001631           \\
        16            & 0.9531        & 0.000903           \\ \hline
        \end{tabular}
    }
    ~
    \subfloat[][TinyImageNet]{
        \begin{tabular}{rrr}
        \hline
        V Factor & Test Accuracy & Std dev. \\ \hline
        2             & 0.5254        & 0.003548           \\
        4             & 0.5193        & 0.003017           \\
        8             & 0.5206        & 0.005651           \\
        16            & 0.5172        & 0.002984           \\ \hline
        \end{tabular}
    }
    \caption{Sensitivity analysis data}
    \label{sens-data}
\end{table}

\section{Dropout Sparsity Learning with the Hard Concrete Distribution}

We describe the Hard Concrete distribution, and how it can be used for Dropout Sparsity Learning.

\subsection{The Hard Concrete Distribution}

\begin{figure}
  \centering
  \includegraphics[width=\textwidth]{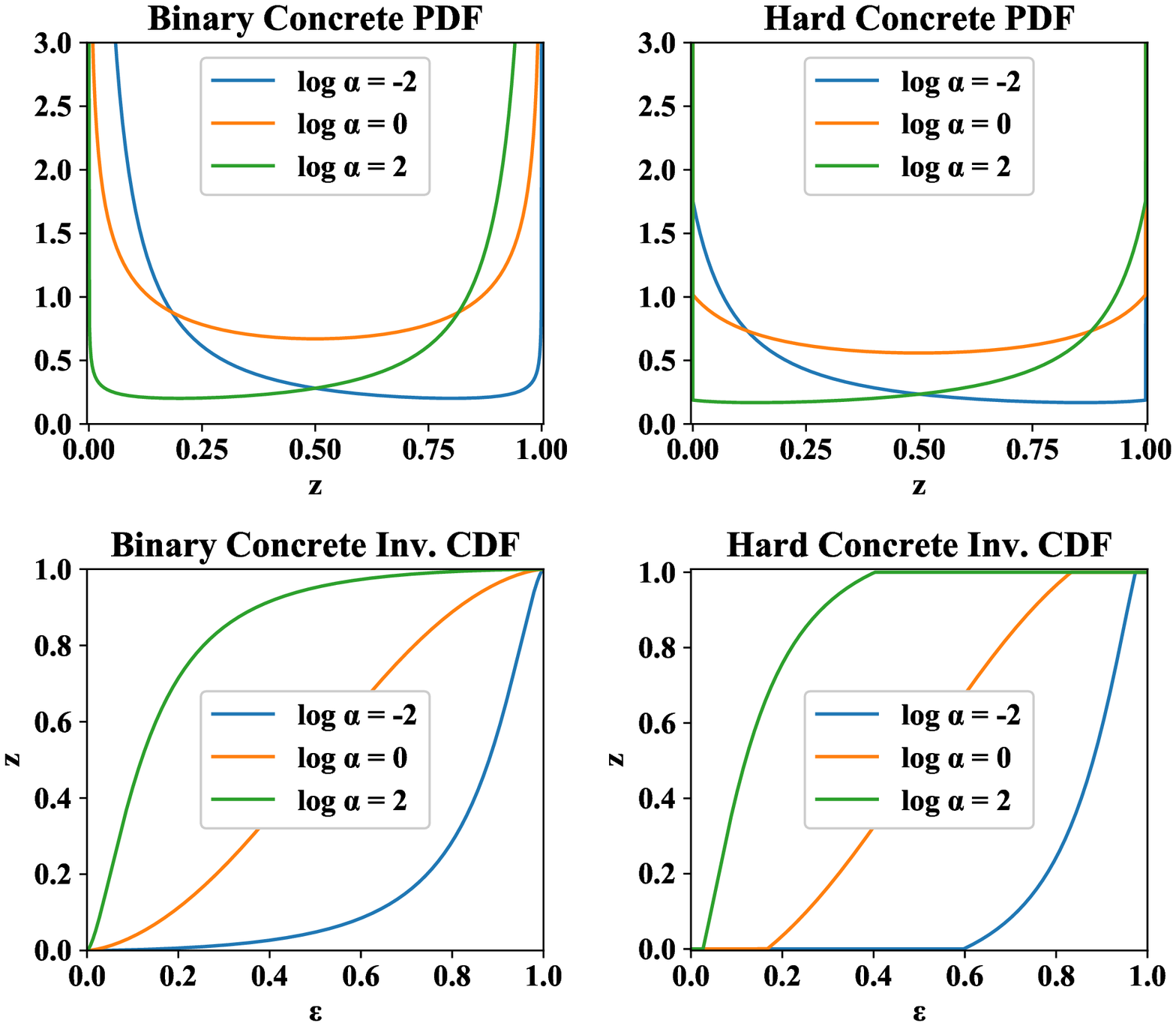}
  \caption{\textbf{Comparison of the BC and the HC distributions}. The parameters $\beta, \gamma, \zeta$ are set to $2/3, -0.1, 1.1$.}
  \label{fig:distributions}
\end{figure}

The Hard Concrete distribution~[22] (HC) is a modification of the Binary Concrete distribution (BC), which in turn is a special case of the Concrete distribution from Maddison et al. (2017). The BC is a continuous relaxation of the Bernouilli distribution. The HC has the advantage of allowing to put significant mass on $P(z=0)$, where $z \in [0,1]$ is the relevance of a neuron (or group of neurons). We are interested in drawing samples $z$ from the HC, which can be done by using $z = Q_\mathrm{HC}^{-1}(\epsilon|\phi)$, where $Q^{-1}$ is the inverse cumulative distribution function (ICDF) and $\epsilon \sim \mathcal{U}(0,1)$. \figref{fig:distributions} compares the probability density functions (PDF) and ICDF of the BC and the HC distributions.

The PDF $q_\mathrm{BC}(s|\phi)$ and CDF $Q_\mathrm{BC}(s|\phi)$ of the BC have parameters $\phi := (\alpha, \beta)$ and are defined as follows:
\begin{align}
    q_\mathrm{BC}(s|\phi) &= \frac{\beta \alpha s^{-\beta-1} (1-s)^{-\beta-1}}{(\alpha s^{-\beta} + (1-s)^{-\beta})^2} \\
    Q_\mathrm{BC}(s|\phi) &= \mathrm{Sigmoid}((\log s - \log(1-s)) \beta - \log \alpha).
\end{align}

To obtain the HC, we stretch the domain of $q_\mathrm{BC}$ to the $(\gamma, \zeta)$ interval, with $\gamma < 0$ and $\zeta > 1$. Since this stretching operation results in some probability mass being outside $[0,1]$, we assign the mass of $[\gamma, 0]$ and $[1, \zeta]$ to $P(z=0)$ and $P(z=1)$, respectively. For example, with $\log \alpha = 0$, $P(z=0)=P(z=1)\approx0.23$. The resulting PDF is better understood visually (c.f. \figref{fig:distributions}). The HC distribution has parameters $\phi := (\alpha, \beta, \gamma, \zeta)$, and we set $\beta, \gamma, \zeta$ to $2/3, -0.1, 1.1$ for all our experiments, as per [22].

For our purposes, we only need to draw samples for the HC, which can be done with its inverse CDF. In our experiments, we use the following formula, given by [22]~:

\begin{equation}
\begin{split}
    Q_\mathrm{HC}^{-1}(\epsilon|\phi) &= \mathrm{Clamp}_{0,1} \left[ \mathrm{Sigmoid} \left( \frac{\log \epsilon - \log(1-\epsilon) + \log \alpha}{\beta} \right) (\zeta - \gamma) + \gamma \right].
\end{split}
\end{equation}

\subsection{The Hard Concrete Sparsity Loss}

We define our sparsity loss as the expectation of the $L_0$ norm of the set of all dropout variables $z$. We replace this discrete norm by a continuous relaxation $L_\mathrm{HC}(\Phi)$, where $\Phi := \{\phi_i\}$ and $\phi := (\alpha, \beta, \gamma, \zeta)$~:

\begin{equation}
    L_\mathrm{HC}(\Phi) = \sum_{\phi \in \Phi} P_\phi(z>0) = \sum_{\phi \in \Phi} \left( 1 - Q_\mathrm{HC} (0|\phi) \right) = \sum_{\phi \in \Phi} \mathrm{Sigmoid}(\log \alpha - \beta \log \frac{-\gamma}{\zeta})
\end{equation}